\documentclass[10pt,twocolumn,letterpaper]{article}

\usepackage{float}
\usepackage[caption = false]{subfig}
\usepackage[final]{graphicx}

\usepackage{cvpr}
\usepackage{times}
\usepackage{epsfig}
\usepackage{graphicx}
\usepackage{amsmath}
\usepackage{amssymb}

\usepackage[breaklinks=true,bookmarks=false]{hyperref}

\cvprfinalcopy 

\def\cvprPaperID{****} 

\begin{document}

\title{CUTIE: Learning to Understand Documents with Convolutional Universal Text Information Extractor}

\author{
  Xiaohui Zhao, Endi Niu, Zhuo Wu, and Xiaoguang Wang \\
  New IT Accenture \\
{\tt\small xh.zhao@outlook.com henryniu14@gmail.com zhuo.wu@accenture.com danny.x.wang@accenture.com}
}

\maketitle

\begin{abstract}
   Extracting key information from documents, such as receipts or invoices, and preserving the interested texts to structured data is crucial in the document-intensive streamline processes of office automation in areas that includes but not limited to accounting, financial, and taxation areas. To avoid designing expert rules for each specific type of document, some published works attempt to tackle the problem by learning a model to explore the semantic context in text sequences based on the Named Entity Recognition (NER) method in the NLP field. In this paper, we propose to harness the effective information from both semantic meaning and spatial distribution of texts in documents. Specifically, our proposed model, Convolutional Universal Text Information Extractor (CUTIE), applies convolutional neural networks on gridded texts where texts are embedded as features with semantical connotations. We further explore the effect of employing different structures of convolutional neural network and propose a fast and portable structure. We demonstrate the effectiveness of the proposed method on a dataset with up to $4,484$ labelled receipts, without any pre-training or post-processing, achieving state of the art performance that is much better than the NER based methods in terms of either speed and accuracy. Experimental results also demonstrate that the proposed CUTIE model being able to achieve good performance with a much smaller amount of training data.
\end{abstract}

\section{Introduction}

Implementing Scanned receipts OCR and information extraction (SROIE) is of great benefit to services and applications such as efficient archiving, compliance check, and fast indexing in the document-intensive streamline processes of office automation in areas that includes but not limited to accounting, financial, and taxation areas. There are two specific tasks involved in SROIE: receipt OCR and key information extraction. In this work, we focus on the second task that is rare in published research. In fact, key information extraction faces big challenges, where different types of document structures and the vast number of potential interested key words introduces great difficulties. Although the commonly used rule-based method can be implemented with carefully designed expert rules, it can only work on certain specific type of documents and takes no lesser effort to adapt to new type of documents. Therefore, it is desirable to have a learning-based key information extraction method with limited requirement of human resources and solely employing the deep learning technique without designing an expert rule for any specific type of documents. 

CloudScan is a learning based invoice analysis system \cite{cloudscan}. Aiming to not rely on invoice layout templates, CloudScan trains a model that could be generalized to unseen invoice layouts, where a model is trained either using Long Short Term Memory (LSTM) or Logistic Regression (LR) with expert designed rules as training features. This turns out to be extremely similar with solving a Named Entity Recognition (NER) or slot filling task. For that reason, several models can be employed, e.g., the Bi-directional Encoder Representations from Transformers (BERT) model is a recently proposed state of the art method and has achieved great success in a wide of range of NLP tasks including NER \cite{bert}. However, the NER models were not originally designed to solve the key information extraction problem in SROIE. To employ NER models, text words in the original document are aligned as a long paragraph based on a line-based rule. In fact, documents, such as receipts and invoices, present with various styles of layouts that were designed for different scenarios or from different enterprise entities. The order or word-to-word distance of the texts in the line-base-aligned long paragraph tend to vary greatly due to layout variations, which is difficult to be handled with the natural language oriented methods. Typical Examples of documents with different layouts are illustrated in Fig. \ref{fig:receipts}.

\begin{figure*}
\begin{center}
\includegraphics[width=0.99\linewidth]{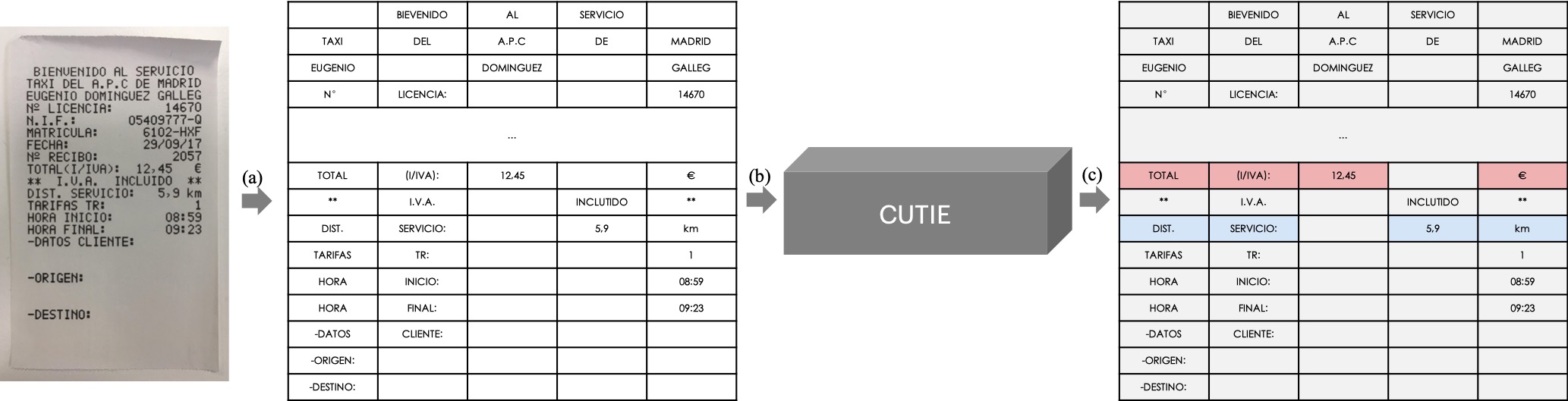}
\end{center}
   \caption{Framework of the proposed method, (a) positional map the scanned document image to a grid with text's relative spatial relation preserved, (b) feed the generated grid into the CNN for extracting key information, (c) reverse map the extracted key information for visual reference.}
\label{fig:cutie}
\end{figure*}
In this work, attempting to involve the spatial information into the key information extraction process, we propose to tackle this problem by using the CNN based network structure and involve the semantic features in a carefully designed fashion. In particular, our proposed model, called Convolutional Universal Text Information Extractor (CUTIE), tackles the key information extraction problem by applying convolutional deep learning model on the gridded texts, as illustrated in Fig. \ref{fig:cutie}. The gridded texts are formed with the proposed grid positional mapping method, where the grid is generated with the principle that is preserving texts’ relative spatial relationship in the original scanned document image. The rich semantic information is encoded from the gridded texts at the very beginning stage of the convolutional neural network with a word embedding layer. The CUTIE allows for simultaneously looking into both semantical information and spatial information of the texts in the scanned document image and can reach a new state of the art result for key information extraction, which outperforms BERT model but without demanding of pretraining on a huge text dataset \cite{bert,transformer}.

\section{Related Works}
Several rule-based invoice analysis systems were proposed in \cite{3,5,7}. Intellix by DocuWare requires a template being annotated with relevant fields \cite{3}. For that reason, a collection of templates have to be constructed. SmartFix employs specifically designed configuration rules for each template \cite{5}. Esser et al. uses a dataset of fixed key information positions for each template \cite{7}. It is not hard to find that the rule-based methods rely heavily on the pre-defined template rules to extract information from specific invoice layouts.

CloudScan is a work attempting to extract key information with learning based models \cite{cloudscan}. Firstly, N-grams features are formed by connecting expert designed rules calculated results on texts of each document line. Then, the features are fed to train a RNN-based or a logistic regression based classifier for key information extraction. Certain post-processings are added to further enhance the extraction results. However, the line-based feature extraction method can not achieve its best performance when document texts are not perfectly aligned. Moreover, the RNN-based classifier, bi-directional LSTM model in CloudScan, has limited ability to learn the relationship among distant words. 

Bidirectional Encoder Representations from Transformers (BERT) is a recently proposed model that is pre-trained on a huge dataset and can be fine-tuned for a specific task, including Named Entity Recognition (NER), which outperforms most of the state of the art results in several NLP tasks \cite{bert}. Since the previous learning based methods treat the key information extraction problem as a NER problem, applying BERT can achieve a better result than the bi-LSTM in CloudScan.

\section{Methods}
In this section, we introduce the method proposed for creating grid data for model training. We then present the network architectures that capture long distance information and avoid information loss in the convolutional neural networks that have striding or pooling processes.

\subsection{Grid Positional Mapping}
\label{pm}
To generate input grid data for the convolutional neural network, the scanned document image are processed by an OCR engine to acquire the texts and their absolute / relative positions. Let the scanned document image be of shape $(w, h)$, the minimum bounding box around the $i$-th interested text $s_i$ be $b_i$ that is restricted by two corner coordinates, where the upper-left corner coordinate in the scanned document be $(x^i_{left}, y^i_{top})$ and the bottom right of the bounding box be $(x^i_{right}, y^i_{bottom})$. To avoid the affects from overlapped bounding boxes and reveal the actual relative position among texts, we calculate the center point $(c^i_x, c^i_y)$ of the bounding boxes as the reference position. It is not hard to find that involving pre-processes that combine texts into meaningful entities will benefit the grid positional mapping process. However, this is not the major purpose of this paper and we leave it to future researches. In this paper, we tokenize the text words with a greedy longest-match-first algorithm using a pre-defined dictionary \cite{bertgit}. 

Let the grid positional mapping process be $G$ and the target grid size be $(c_{g_m}, r_{g_m})$. To generate the grid data, the goal of $G$ is to map the texts from the original scanned document image to the target grid, such that the mapped grid preserves the original spatial relationship among texts yet more suitable to be used as the input for the convolutional neural network. The mapping position of texts in the grid is calculated as

\begin{equation}
\label{cx}
c^i_x = c_{g_m} \frac{x_{left} + \frac{(x_{right} - x_{left})}{2}}{w}
\end{equation}
\begin{equation}
\label{cy}
r^i_y = r_{g_m} \frac{y_{top} + \frac{(y_{bottom} - y_{top})}{2}}{h}
\end{equation}

For tokenized texts, the bounding box is horizontally divided into multiple boxes and their row and col reference positions are calculated using the same criteria as Equ. \ref{cx} and Equ. \ref{cy}, separately. Furthermore, to enhance the capability of CUTIE to better handle documents with different layouts, we augment the grid data to shapes with different rows and columns by random sampling a Gaussian distribution with probability

\begin{equation}
\label{augmentc}
p_c(k) = \frac{1}{\sqrt{2 \pi \sigma^2}} e^{- \frac{(k - c_{g_t})^2}{2 \sigma^2}}
\end{equation}
\begin{equation}
\label{augmentr}
p_r(k) = \frac{1}{\sqrt{2 \pi \sigma^2}} e^{- \frac{(k - r_{g_t})^2}{2 \sigma^2}}
\end{equation}
where $c_{g_t}$ is the mean center of the target augment gird size, $r_{g_t}$ is the mean center of the target augment grid size, and $\sigma$ is the standard deviation. In case of two tokens occupy the same grid cell, tokens in the same row are shifted to place the tokens. This is acceptable since the model learns the relative spatial relationship among tokens and convolutional neural networks handle translations well.

\begin{figure}
\begin{center}
\includegraphics[width=0.99\linewidth]{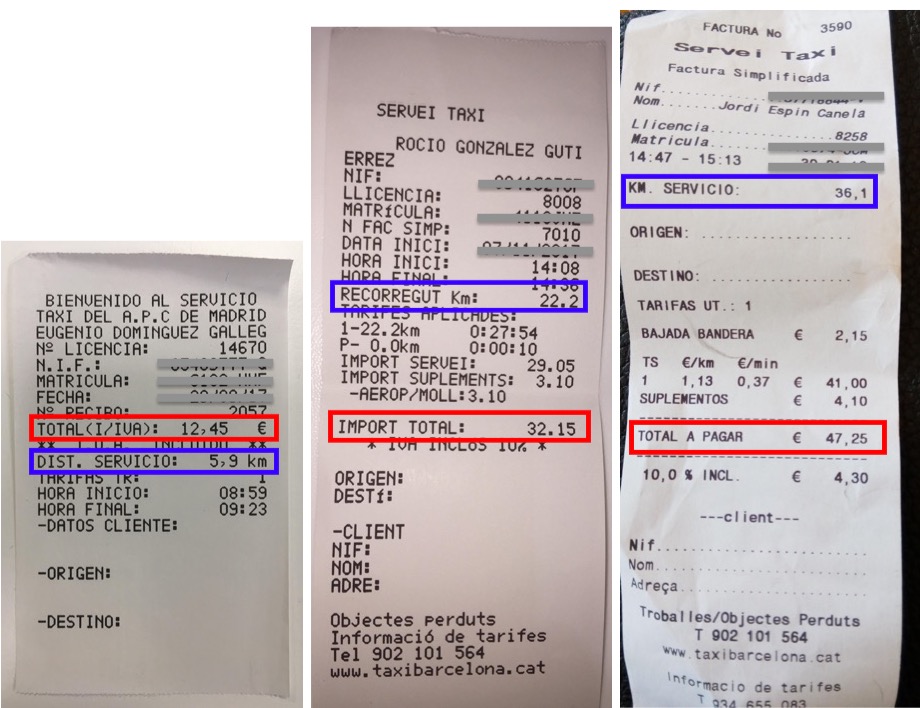}
\end{center}
   \caption{Example of scanned taxi receipt images. We provide two colored rectangles to help readers find the key information about distance of travel and total amount with blue and red, respectively. Note the different types of spatial layouts and key information texts in these receipt images. }
\label{fig:receipts}
\end{figure}

\subsection{CUTIE Model}
Through matching the output of CUTIE with the labelled grid data, the model learns to generate the label for each text in the grid input via exploring both the spatial and semantic features. For that reason, the task of CUTIE bears resemblance to the semantic segmentation task in the computer vision field but with more sparse data distributions. Specifically, the mapped grid contains scattered data points (text tokens) in contrast to the images bespread with pixels. The grid positional mapped key texts are either close to or distant to each other due to different types of document layouts. Therefore, incorporating multi-scale context processing ability benefits the network.

In fact, several methods have been proposed in the semantic segmentation field to capture multi-scale contexts in the input data. The methods of image pyramid and the encoder-decoder structure both aim at exploiting multi-scale information. The interested objects from different scales become prominent in the former networks by using multiple scaled input data to gather multi-scale features. The later networks shrink feature maps to enlarge receptive fields and reduce computation burdens, and then capture finer details by gradually recovering the spatial information from lower layer features. However, spatial resolution is reduced in the encoding process and the decoding process exploits only high resolution but low-level features to recover the spatial resolution, the consecutive striding encoding process decimates detail information \cite{hrnet}. Moreover, the encoding and decoding process applies shape restricts to the grid shape augmentation process as introduced in Section \ref{pm}. 

Instead, the field of view of filters can also be effectively enlarged and multi-scale contexts can be captured by combining multi-resolution features \cite{hrnet} or by applying atrous convolution \cite{deeplab, deeplabv1, deeplabv3, deeplabv3p}. To capture long distance connection and avoid potential information loss in the encoding process, we propose two different network architectures and compare their performance in Section \ref{experiment}. In fact, we had experimented with various types of model structures and only detail two of them here to avoid being a tedious paper. Specifically, the proposed CUTIE-A is a high capacity convolutional neural network that fuses multi-resolution features without losing high-resolution features, the proposed CUTIE-B is a convolutional network with atrous convolution for enlarging the field of view and Atrous Spatial Pyramid Pooling (ASPP) module to capture multi-scale contexts. 

Both CUTIE-A and CUITE-B conducts semantical meaning encoding process with a word embedding layer in the very beginning stage. Dropout is applied on the embedding layer to enhance the CUTIE's generalization ability. The cross entropy loss function is applied to compare the predicted token class grid and the ground truth grid.

\subsubsection{CUTIE-A}
CUTIE-A avoids information loss in the encoding process while taking advantage of encoders by combining encoding results to the maintained high-resolution representations through the entire convolutional process. Similar to HRNet proposed in \cite{hrnet}, a high-resolution network without striding is employed as the backbone network and several high-to-low resolution sub networks are gradually added and connected to the backbone major network. During the connecting process of the major network and sub networks, multi-scale features are fused to generate rich representations.

\subsubsection{CUTIE-B}
CUTIE-B is constructed with a single backbone network but employs atrous convolution to capture long distance connections. For atrous convolution, let the input feature map be $m$, filter be $w$ and output be $n$, for each position $i$, atrous convolution is applied over the input feature map $m$ as 

\begin{equation}
n[i] = \sum_k m[i+r\cdot k]w[k]
\end{equation}
where $r$ is the atrous rate that indicates the sampling stride of the input signal, which is implemented as convolving the input feature with upsampled filters by inserting $r-1$ zeros between two consecutive filter values along each spatial dimension. Standard convolution is a special case of atrous convolution with $r=1$ \cite{deeplabv1}.

\begin{table}
	\caption{Statistic of the numbers of labelled receipt document images and key information classes in the dataset.}
\begin{center}
\begin{tabular}{l | c | c | c}
	 & Training Set & Test Set & \#classes \\
	\hline
	ME & 1109 & 375 & 9 \\
	Taxi & 1125 & 375 & 6 \\
	Hotel & 1125 & 375 & 9 \\
\end{tabular}
\end{center}
	\label{tab:dataset}
\end{table}

\begin{table*}
	\caption{Performance comparison on different types of documents. (AP/softAP)}
\begin{center}
\begin{tabular}{l | c | c | c | c}
	Method & \#Params & Taxi & ME & Hotel \\
	\hline
	CloudScan\cite{cloudscan} & - & 82 / - & 64 / - & 60 / - \\
	BERT for NER\cite{bert} & 110M & 88.1 / - & 80.1 / - & 71.7 / - \\
	CUTIE-A & 67M & 90.8 / 97.2 & 77.7 / \textbf{91.4} & 69.5 / \textbf{87.8} \\
	CUTIE-B & 14M & \textbf{94.0} / \textbf{97.3} & \textbf{81.5} / 89.7 & \textbf{74.6} / 87.0 \\
\end{tabular}
\end{center}
	\label{tab:comparison}
\end{table*}

\section{Experiments}
\label{experiment}
The proposed method is evaluated on ICDAR 2019 robust reading challenge on SROIE dataset and also on a self-built dataset with $3$ types of scanned document images, which contain $8$ key information classes and $1$ don't care class. For each specific key information class, multiple tokens can be included. The overall performance is referred as strict average precision (AP) and measured in terms of per-class accuracy across the $9$ classes, where one class is determined as correct only when every single token in the class is correct. To achieve deeper analysis of the performance of the proposed method, we propose to use one more criteria, soft average precision (softAP), where the prediction of a key information class is determined as correct as if positive ground truths are correctly predicted even if some false positives are included in the final prediction. SoftAP is important since it indicates model's capability of extracting correct key information with tolerance of incorporating certain false positives. In fact, post processings can be employed to eliminate the false positives. Therefore, joint analysis of AP and softAP provides a better understanding of the model performance.

We compare the performance of the proposed method with two state of the art methods CloudScan \cite{cloudscan} and BERT for NER \cite{bert}. For comparison, the Cloud Scan model for SROIE is trained from scratch but with several expert designed features as described in \cite{cloudscan}. The BERT model for SROIE is transform learned with the Google released base model that is pre-trained on a huge dataset with $3,300$M words \cite{bert,bertgit}. To provide a fair comparison, $4,484$ samples with around $1,500$ each from Taxi, Meals Entertainment (ME), and Hotel, where $3,375$ samples are used for training and $1,125$ are used for testing, as reported in Table \ref{tab:dataset}. A single model is trained on the dataset for all these $3$ types of document receipts either for CloudScan, BERT, or CUTIE.

We use a learning rate of $1e-3$ with Adam optimizer and step decay learning strategy. The learning rate is dropped to $1e-4$ and $1e-5$ on the $15,000$-th and $30,000$-th steps, respectively. The training is terminated within $40,000$ steps with batch size of $32$. We train our model on Tesla V100 GPU where $11$ to $19$GB memories is used depending on the configuration of the model framework and size of the dataset. Instance normalization and hard negative mining are involved to facilate training. Dropout is applied with keep probability of $0.9$. Our model is trained end-to-end without piecewise pretraining of any component. The default embedding size is $128$, target augmentation shape is $64$ for both row and column. The dataset is split as training set and test set with ratio of $75:25$. Neither pre-processing nor post-processing is involved in CUTIE.

\begin{figure*}
\begin{center}
\subfloat{\includegraphics[width=0.17\linewidth]{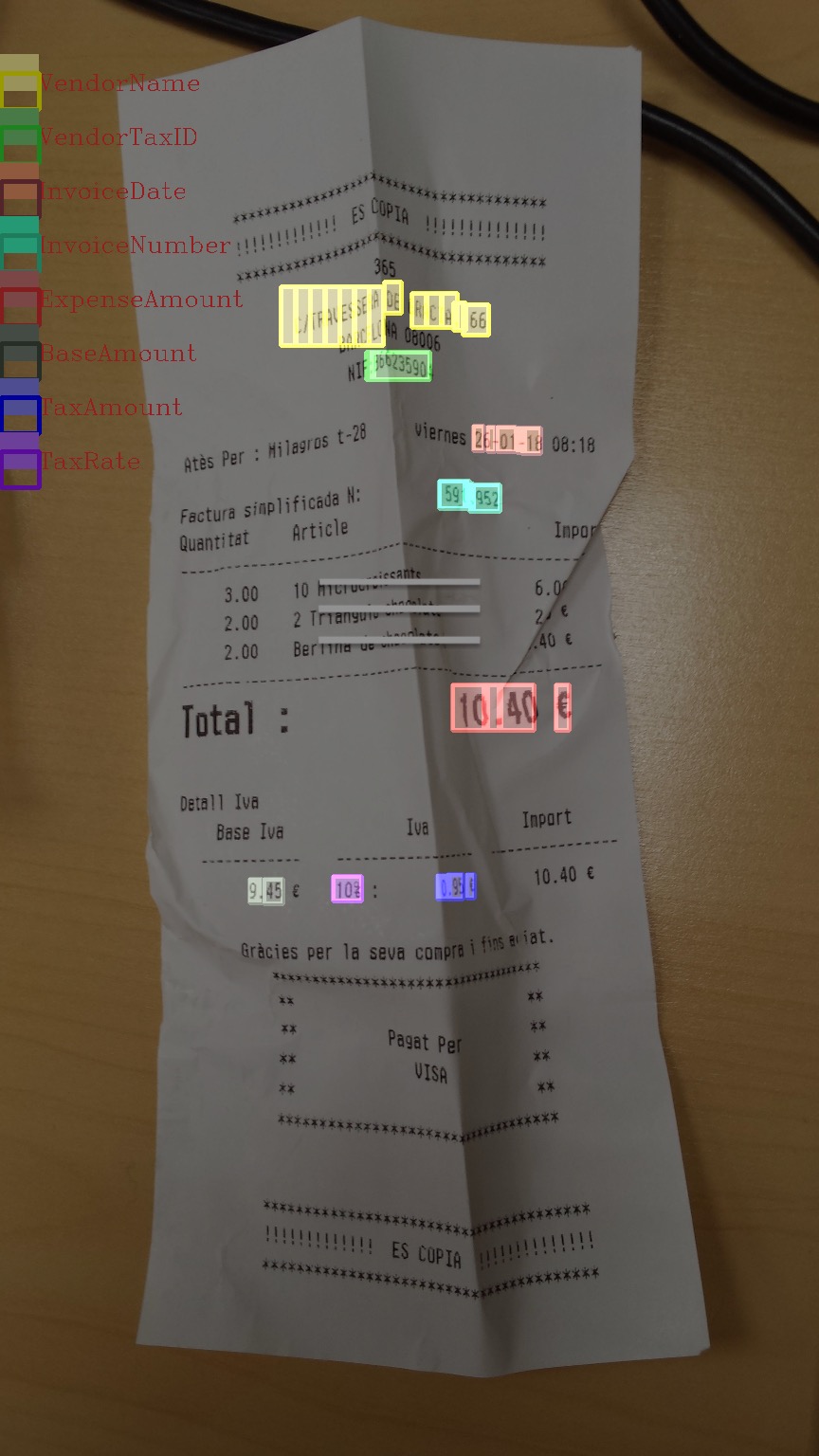}} 
\subfloat{\fcolorbox{white}{white}{}}
\subfloat{\includegraphics[width=0.39\linewidth]{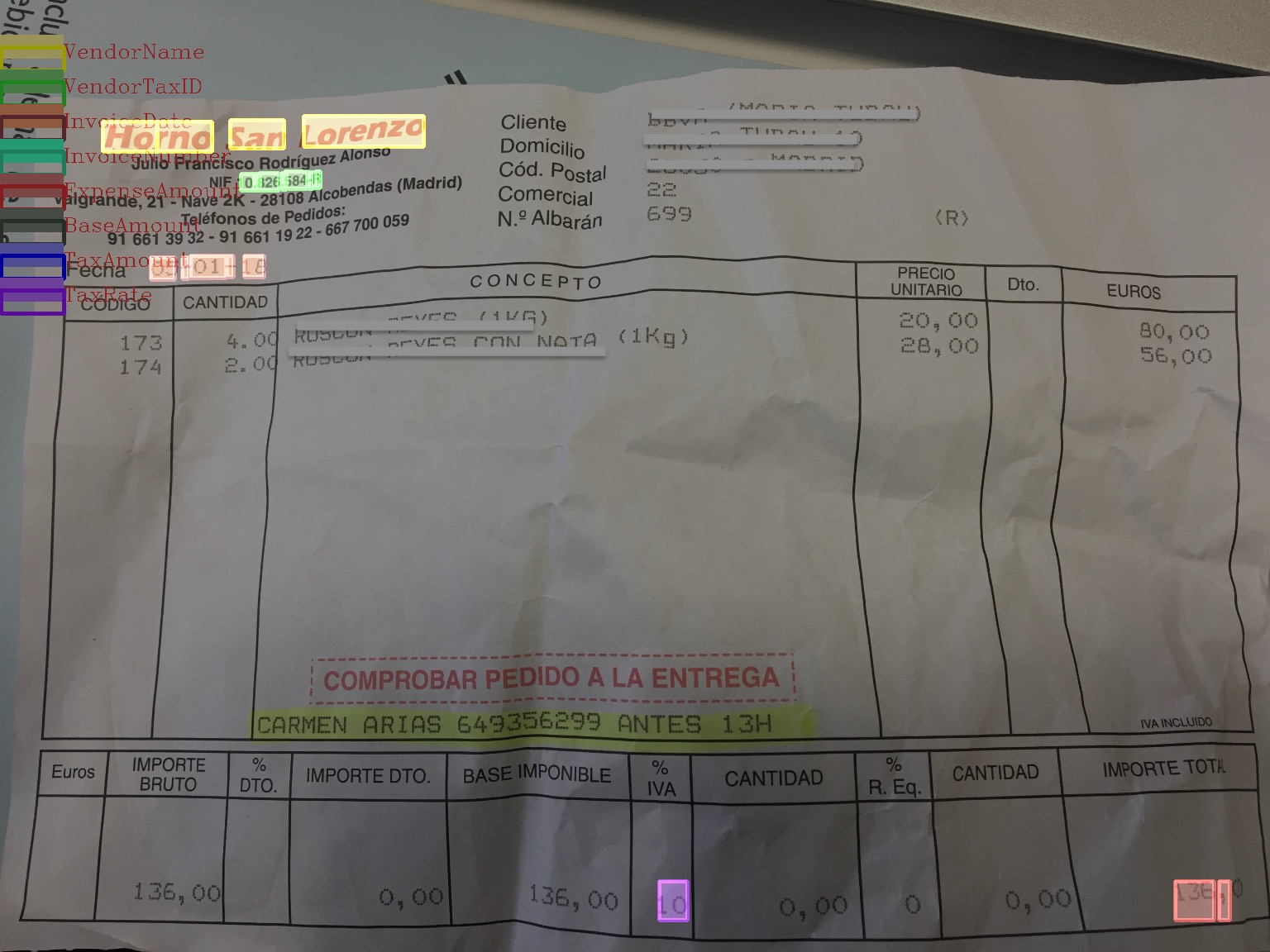}}
\subfloat{\fcolorbox{white}{white}{}}
\subfloat{\includegraphics[width=0.17\linewidth]{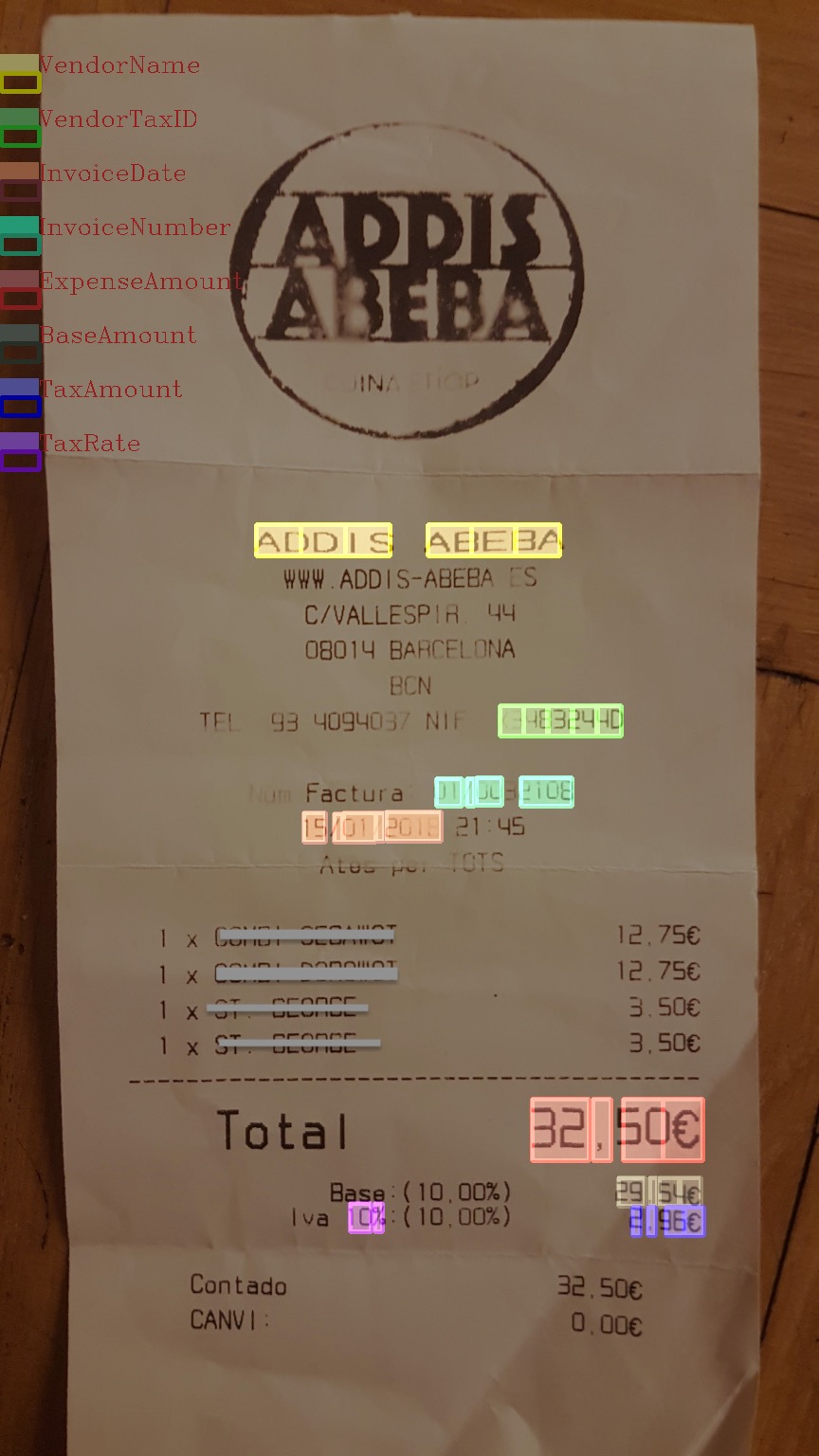}}
\subfloat{\fcolorbox{white}{white}{}}
\subfloat{\includegraphics[width=0.17\linewidth]{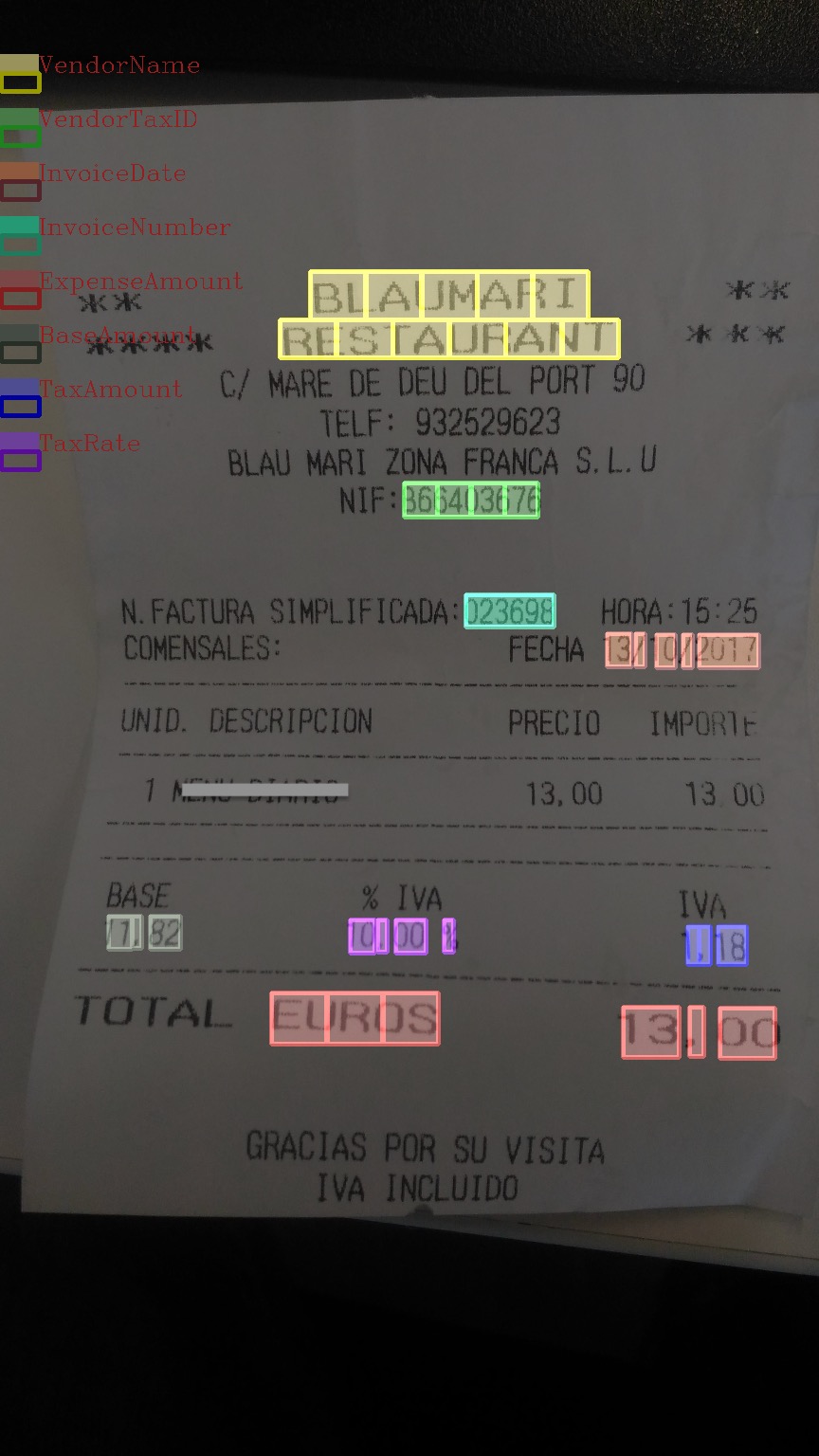}} \\ 
\subfloat{\includegraphics[width=0.19\linewidth]{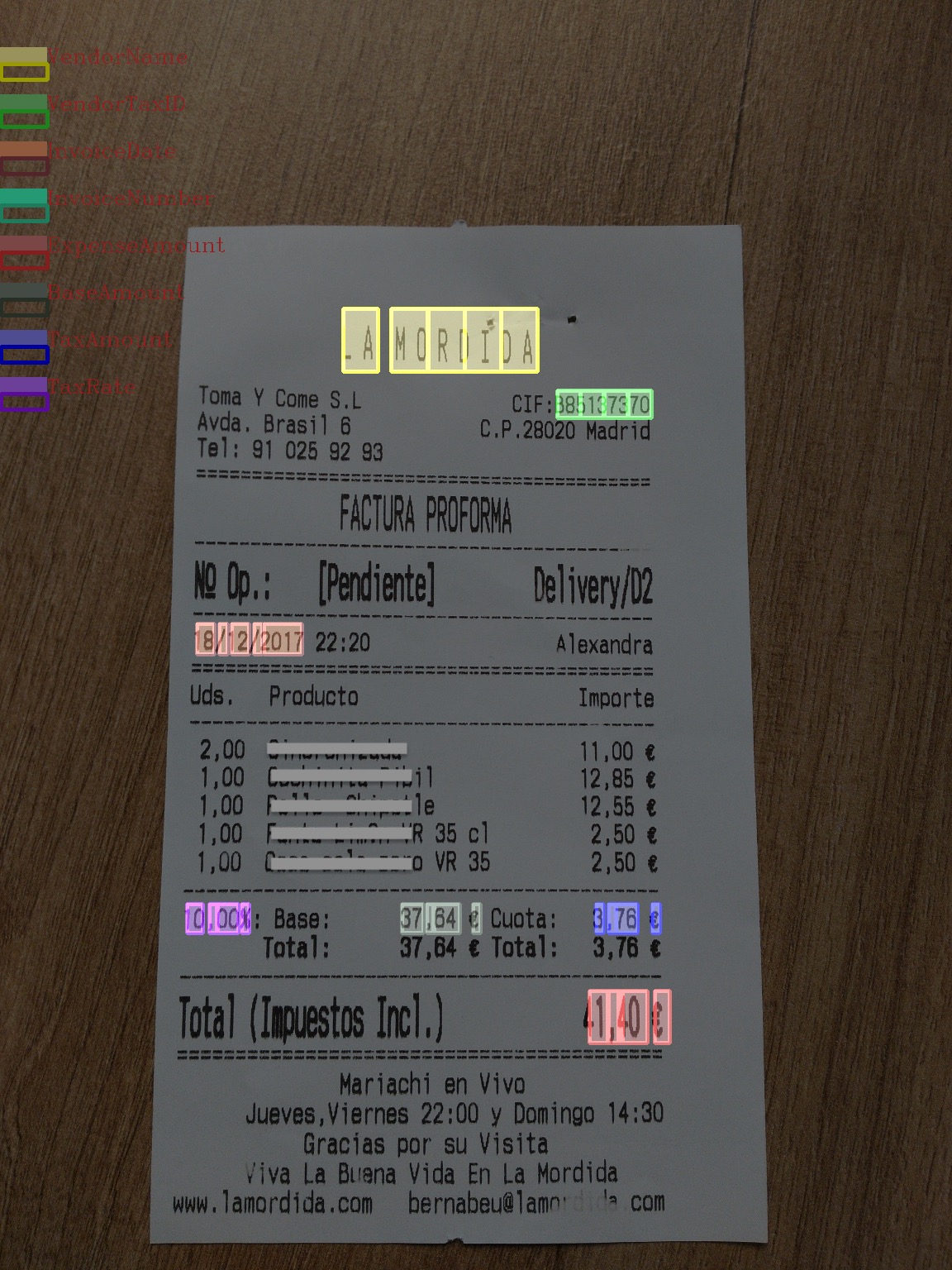}}  
\subfloat{\fcolorbox{white}{white}{}}
\subfloat{\includegraphics[width=0.19\linewidth]{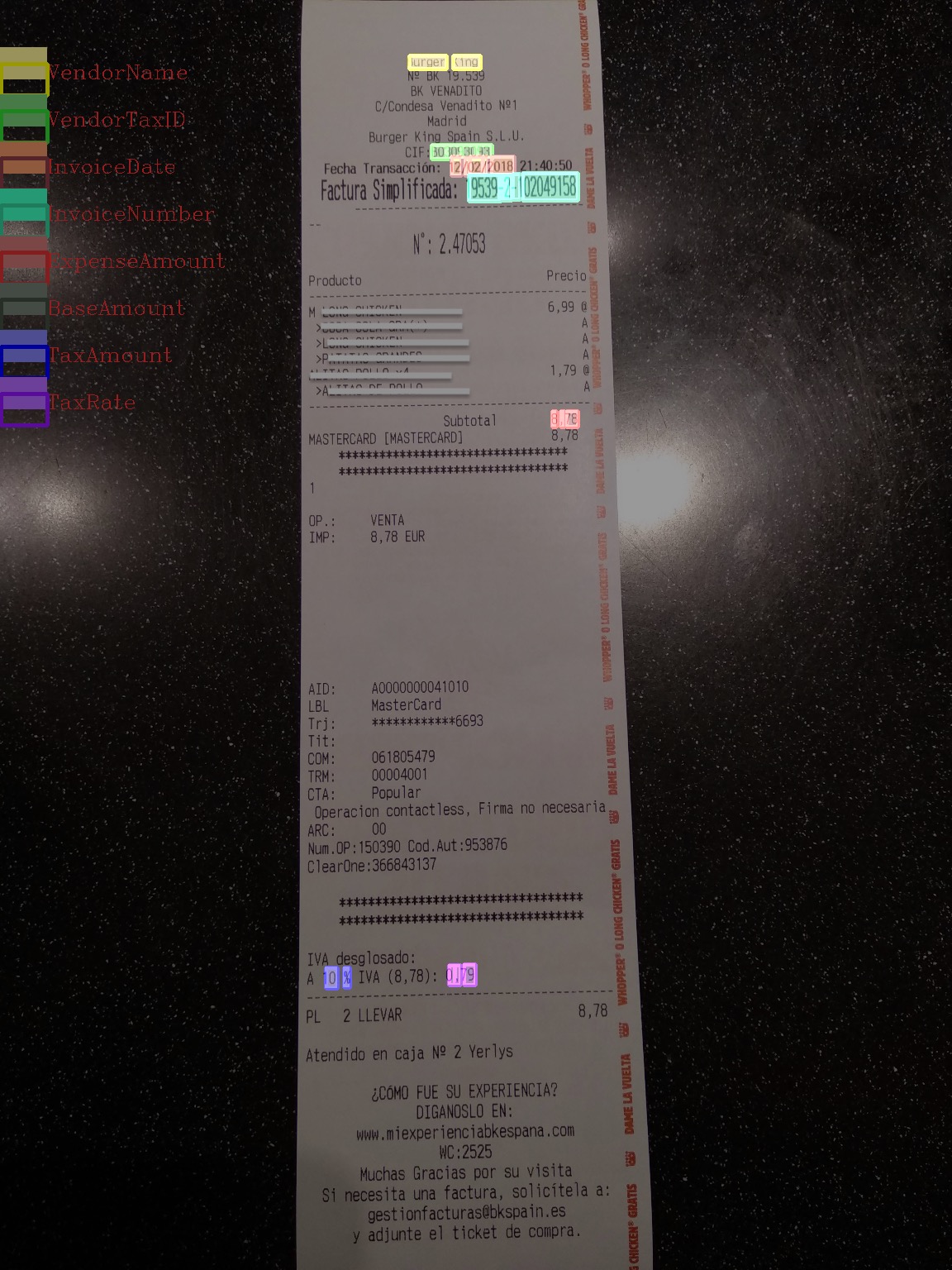}}  
\subfloat{\fcolorbox{white}{white}{}}
\subfloat{\includegraphics[width=0.19\linewidth]{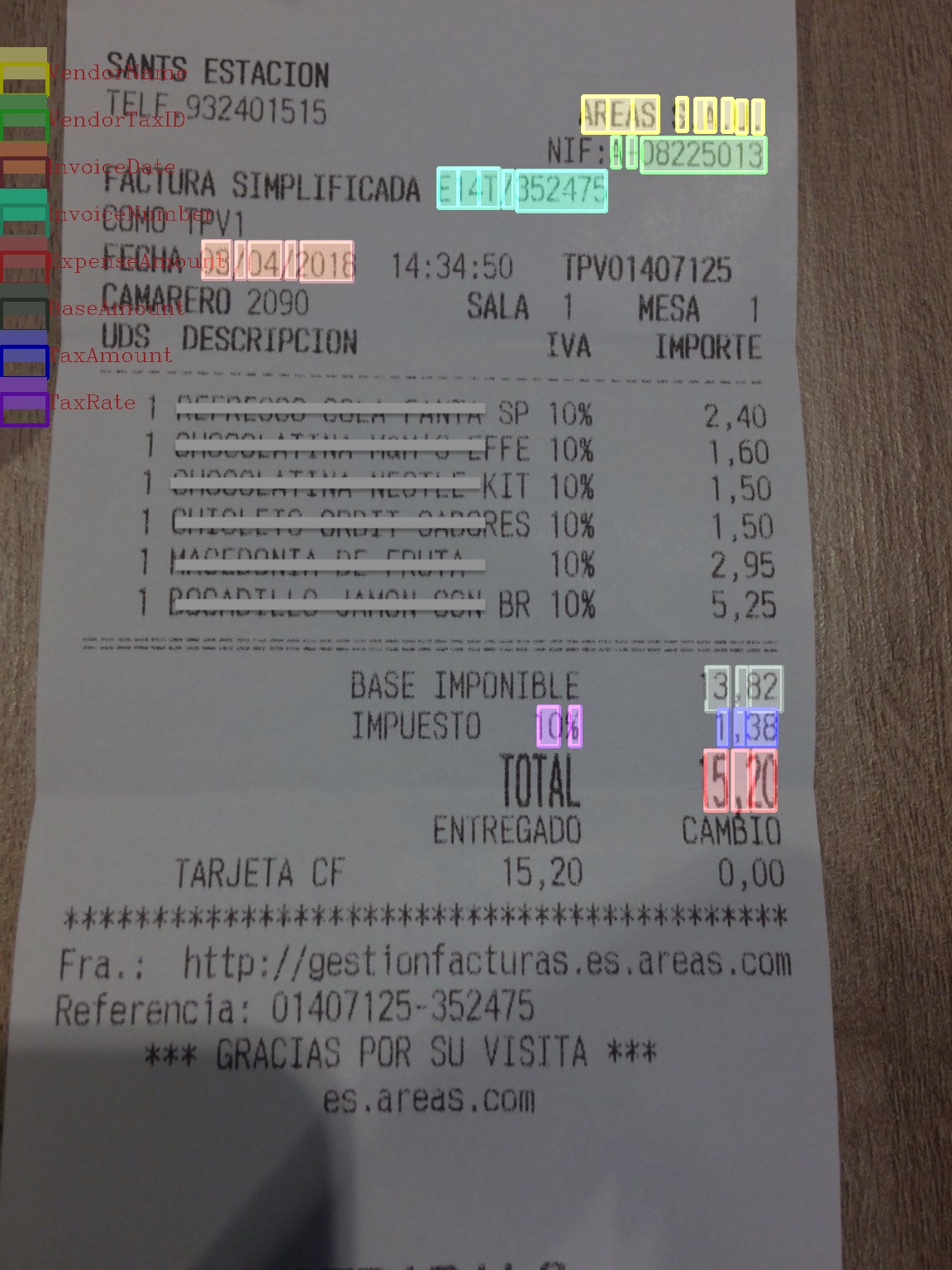}}  
\subfloat{\fcolorbox{white}{white}{}}
\subfloat{\includegraphics[width=0.19\linewidth]{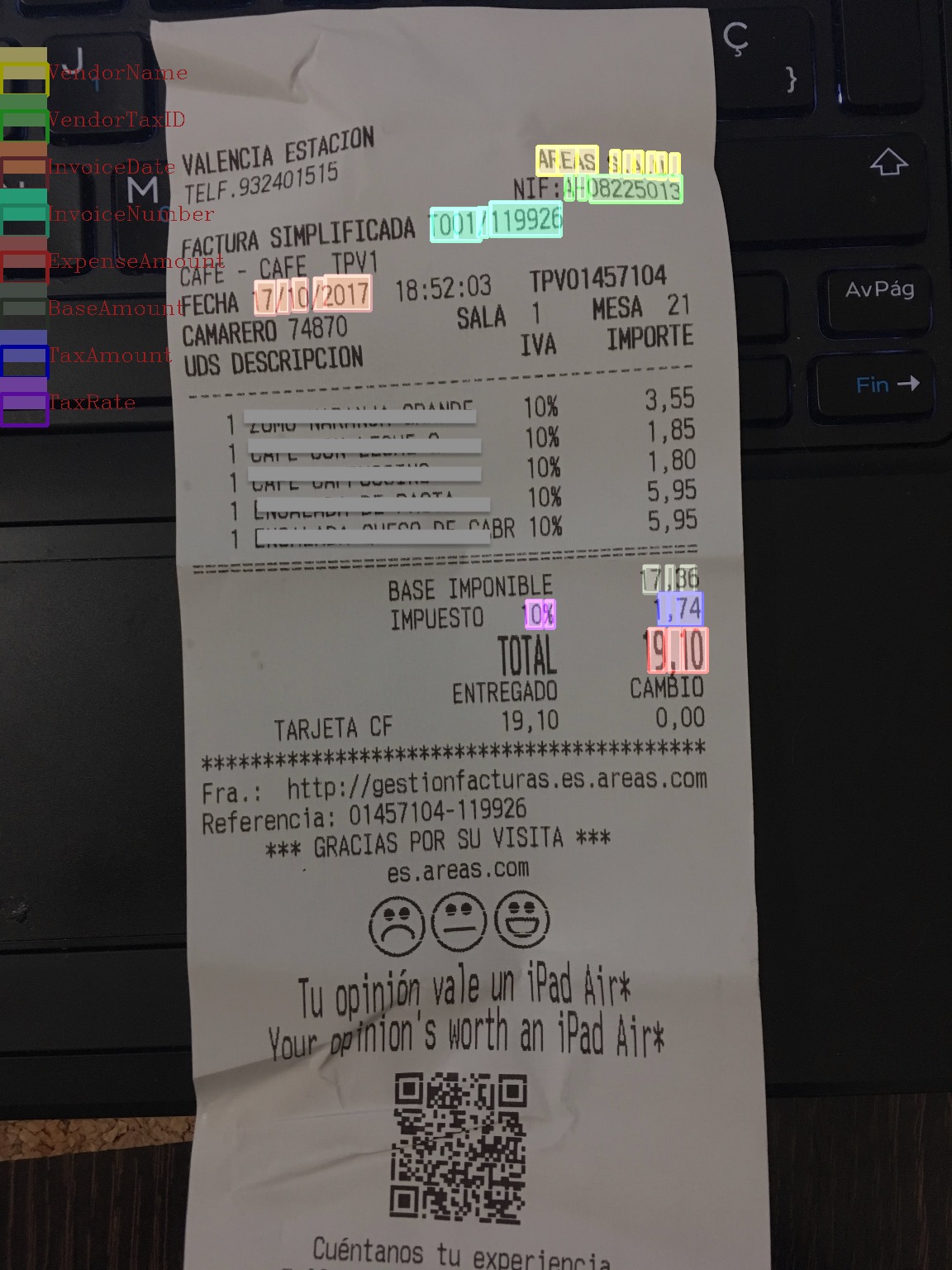}}  
\subfloat{\fcolorbox{white}{white}{}}
\subfloat{\includegraphics[width=0.19\linewidth]{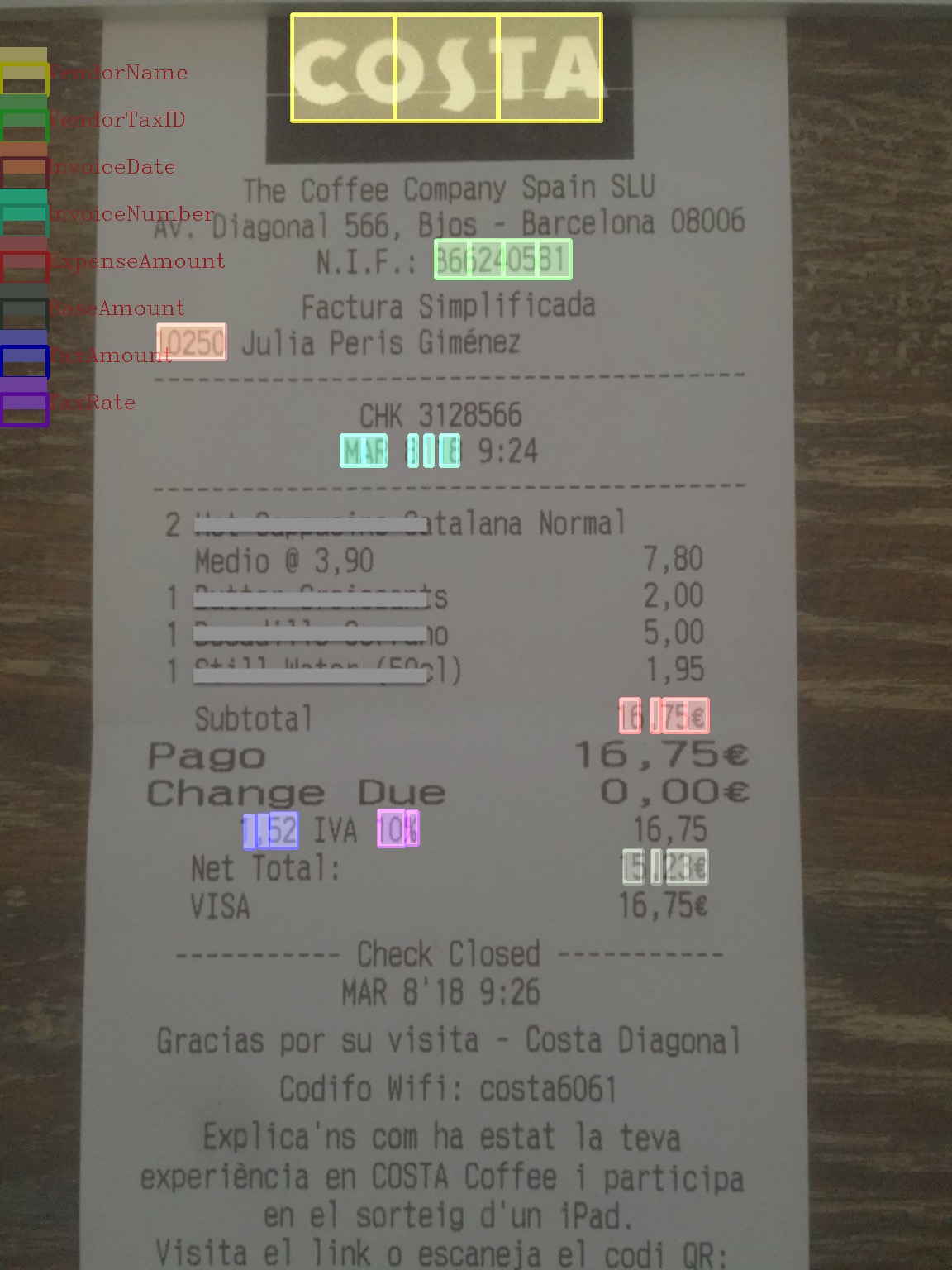}} \\
\subfloat{\includegraphics[width=0.35\linewidth]{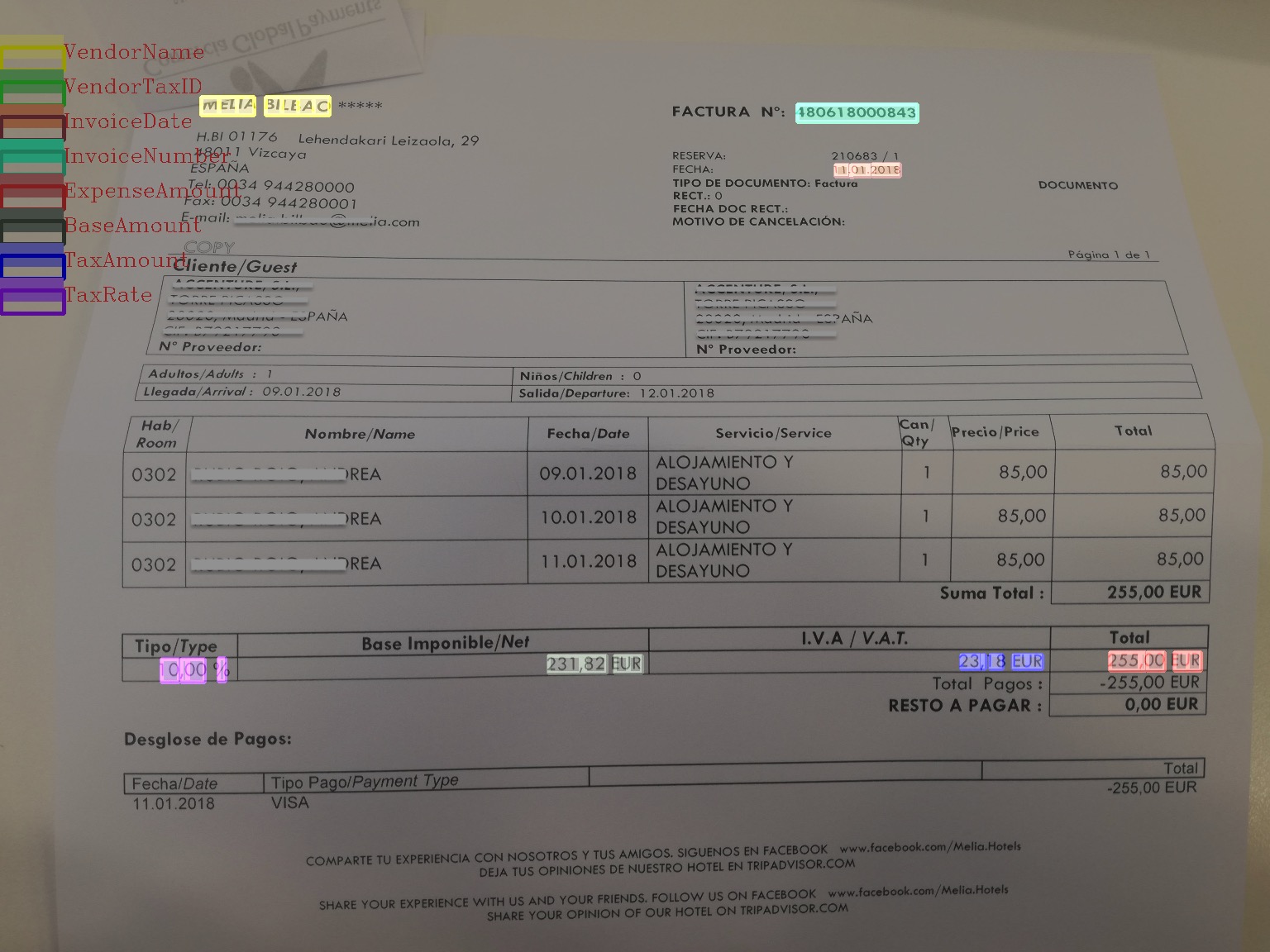}}  
\subfloat{\fcolorbox{white}{white}{}}
\subfloat{\includegraphics[width=0.19\linewidth]{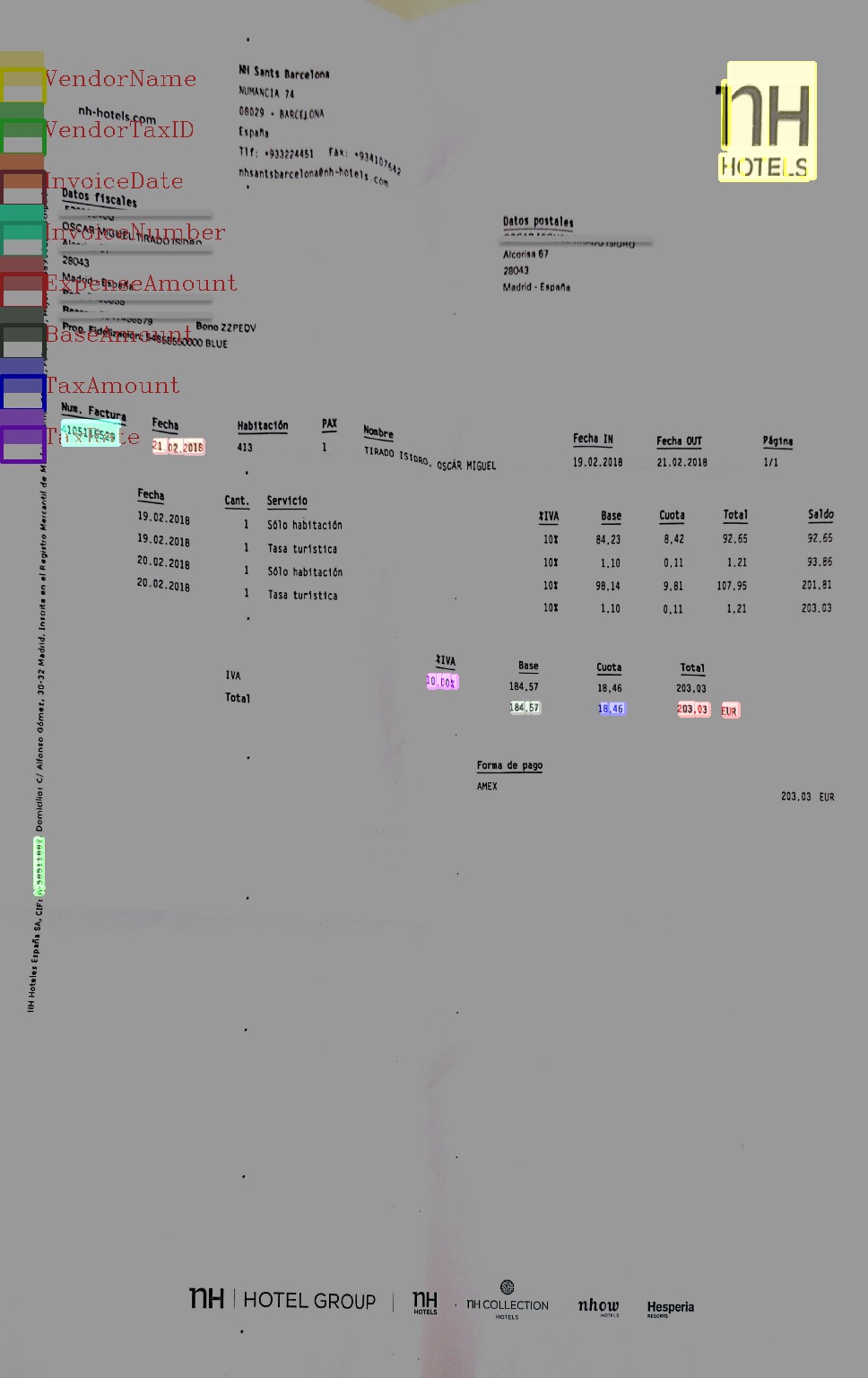}}  
\subfloat{\fcolorbox{white}{white}{}}
\subfloat{\includegraphics[width=0.21\linewidth]{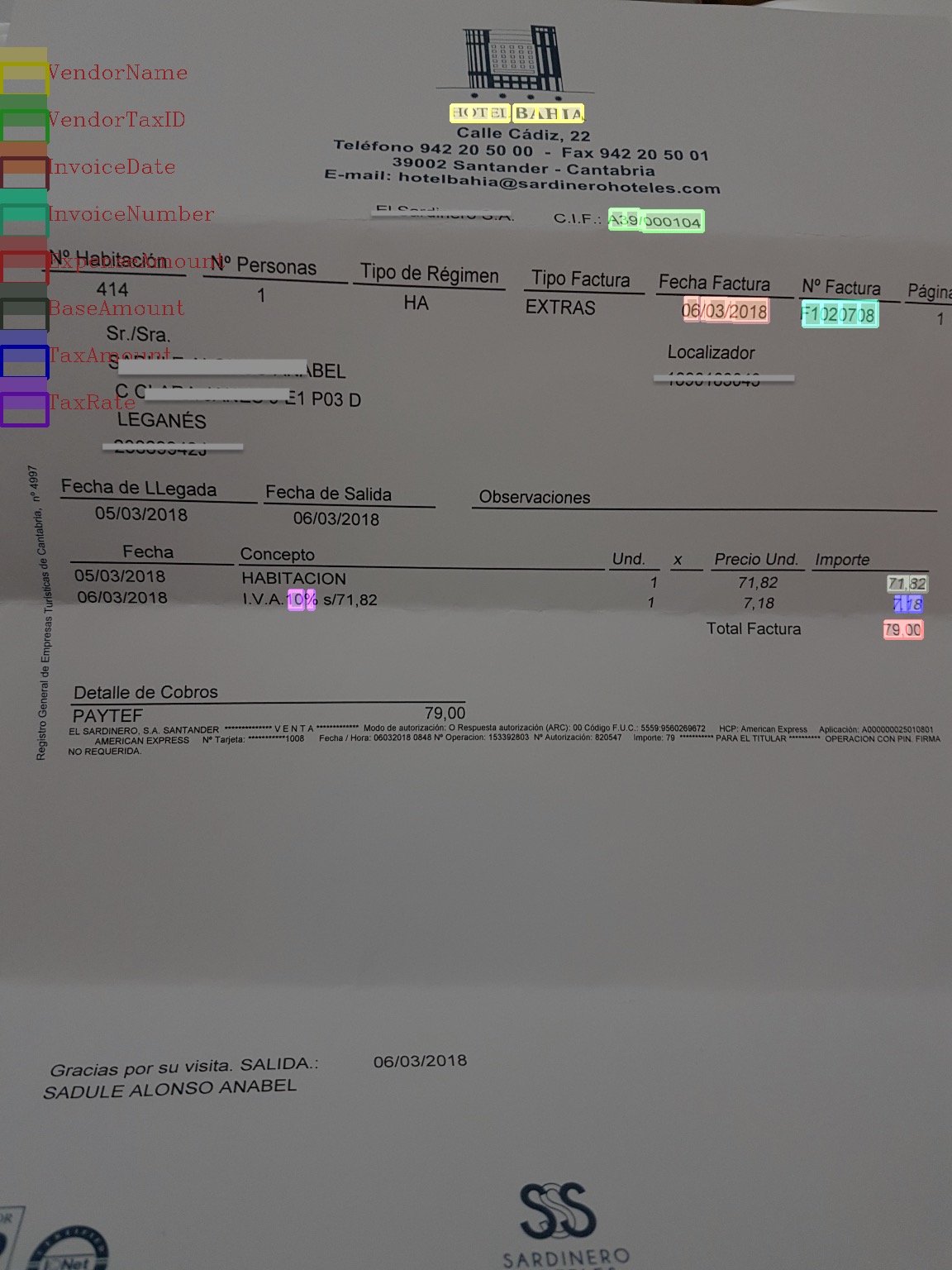}}  
\subfloat{\fcolorbox{white}{white}{}}
\subfloat{\includegraphics[width=0.21\linewidth]{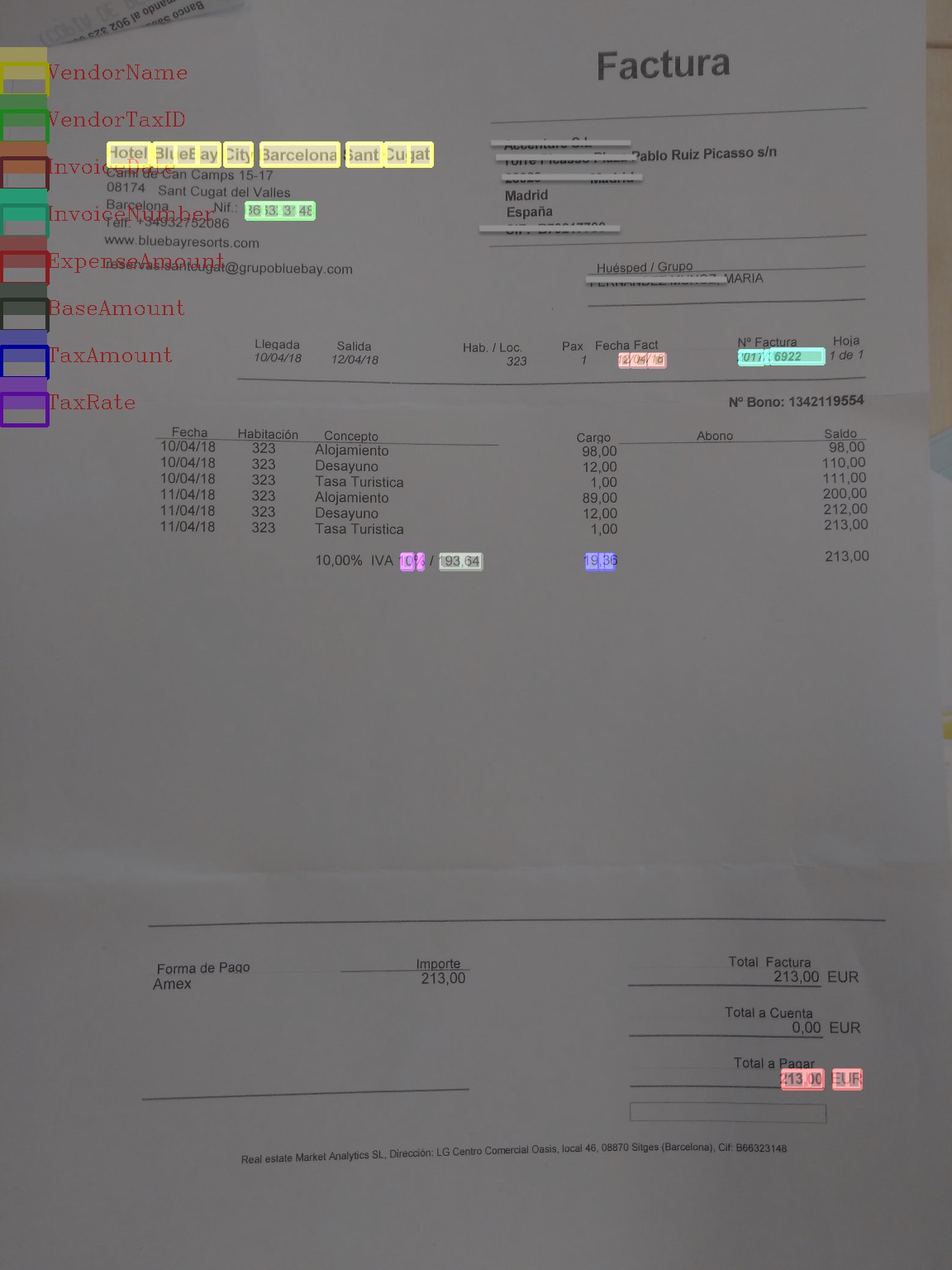}}  
\end{center}
   \caption{Example of CUITE inference results. Color legend in the top-left corner indicates the key information classes. Each color indicates a key information class, where filled rectangles are the ground truths while the boundary-only rectangles are the inference results. The result is perfectly correct as if the filled rectangles overlap with the boundary-only rectangles. We mask out certain private information with filled gray rectangles in these figures. (zooming in to check the details)}
\label{fig:result}
\end{figure*}

\subsection{Dataset}
The ICDAR 2019 SROIE data set (task 3) contains $1000$ whole scanned receipt images. Each receipt image contains around about four key text fields, such as goods name, unit price, date, and total cost. The text annotated in the dataset mainly consists of digits and English characters. The dataset is split into a training/validation set (“trainval”) and a test set (“test”). The “trainval” set consists of $627$ receipt images which is made available to the participants along with their annotations. Since the test annotation is not avaliable yet, we conduct laundry on the trainval set and selected $517$ samples, where the filtered $55$ samples were wrongly labeled. Then we split the trainval set in the ratio of $75:25$, where $75\%$ are used as training data and $25\%$ are used for validation.

The self-built dataset contains $4,484$ annotated scanned Spanish receipt documents, including taxi receipts, meals entertainment (ME) receipts, and hotel receipts, with $9$ different key information classes. We generate the texts and corresponding bounding boxes with Google's OCR API. Each text and their bounding box is manually labelled as one of the $9$ different classes: 'DontCare', 'VendorName', 'VendorTaxID', 'InvoiceDate', 'InvoiceNumber', 'ExpenseAmount', 'BaseAmount', 'TaxAmount', and 'TaxRate'. We then employ the tokenizer introduced in Section \ref{pm} to segment texts into minimum token units, where text bounding boxes are also segmented accordingly. 

The dataset in this work is much more difficult than the neat scanned document images, since various layouts of receipts were captured in different scenarios with mobile phones. Examples of the scanned document images in our dataset are illustrated in Figure \ref{fig:receipts}, note that the colored rectangles are only for visual reference and the actual labelled data are in token-level rather than the line-level as shown in the figure.

\subsection{Overall Performance}
We report results of our method in terms of AP and compare with other state of the art methods in Table \ref{tab:comparison}. We also provide softAP results for CUTIE-A and CUTIE-B in Table \ref{tab:comparison}, where both the softAP of CUTIE-A and CUTIE-B exceeds their AP by a large margin.  Examples of inference results are illustrated in Figure \ref{fig:result}. Our big network CUTIE-A achieves $90.8\%$ AP and $97.2\%$ softAP on taxi receipts, $77.7\%$ AP and $91.4\%$ softAP on meals entertainment receipts, and $69.5\%$ AP and $87.8\%$ softAP on hotel receipts. Compared to CloudScan, CUTIE-A and CUTIE-B outperforms in every single test case. Furthermore, compared to BERT for NER, which is pre-trained on a dataset with $3,300$M words and transfer learned on our dataset, our big network CUTIE-A improves AP by $2.7\%$ AP on taxi receipts but less accurate on other type of documents while using only $1/2$ parameters; our small network CUTIE-B improves AP by $5.9\%$ on taxi receipts, $1.4\%$ points on meals entertainment receipts, and $2.9\%$ on hotel receipts, outperforming other methods in all of the test cases but with much less complexity and smaller model size with only $1/9$ parameters and without requiring a huge dataset for model pretraining. We will further prove in Section \ref{parameters} that CUTIE-B is also able to achieve state of the art performance with only $1/10$ of BERT's parameters. Although CUTIE-B is smaller in capacity, it outperforms CUTIE-A in several assessment criterias. This is because CUTIE-B enlarges the field of view by employing the atrous convolution rather than the pooling or striding processes, the CUTIE-B model has a larger field of view and better understanding of tokens' relative spatial relationship since no restriction is applied on the feature maps shapes. 

It is worth noting that the difference of AP and softAP leads to interesting findings. One of the findings is that CUTIE is capable of extracting the interested texts but sometimes involves texts that are not in the ground truth. Another finding is that the hotel receipts are quite different from the meals entertainment receipts and the taxi receipts, where key information appears multi times in different areas of the receipt, whereas the human labelers tend to label only one of their appearances. We look deeper into this in the following part of this section by analyzing some inference result cases.

\begin{figure}
\begin{center}
\subfloat{\includegraphics[width=0.31\linewidth]{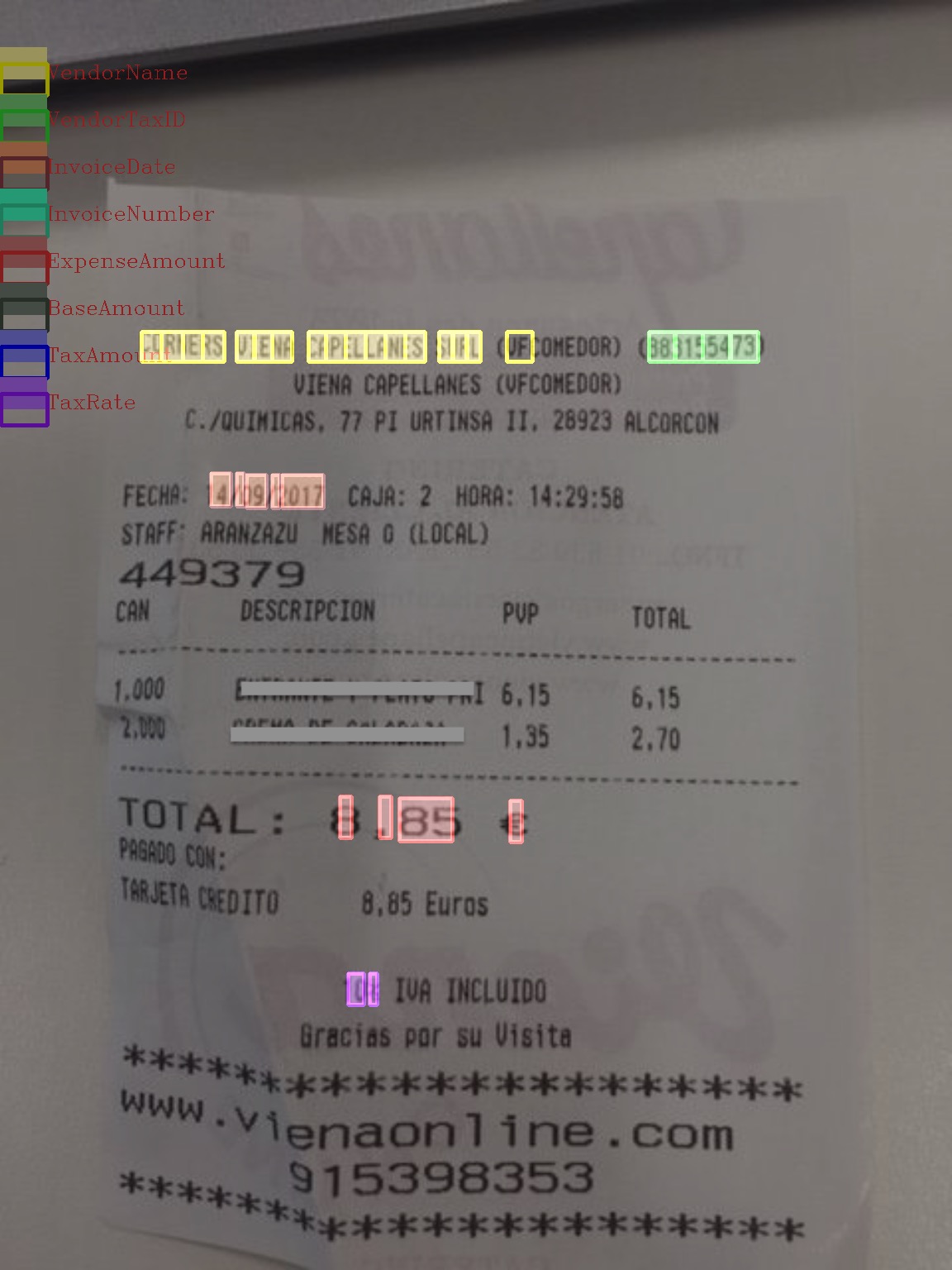}} 
\subfloat{\fcolorbox{white}{white}{}}
\subfloat{\includegraphics[width=0.31\linewidth]{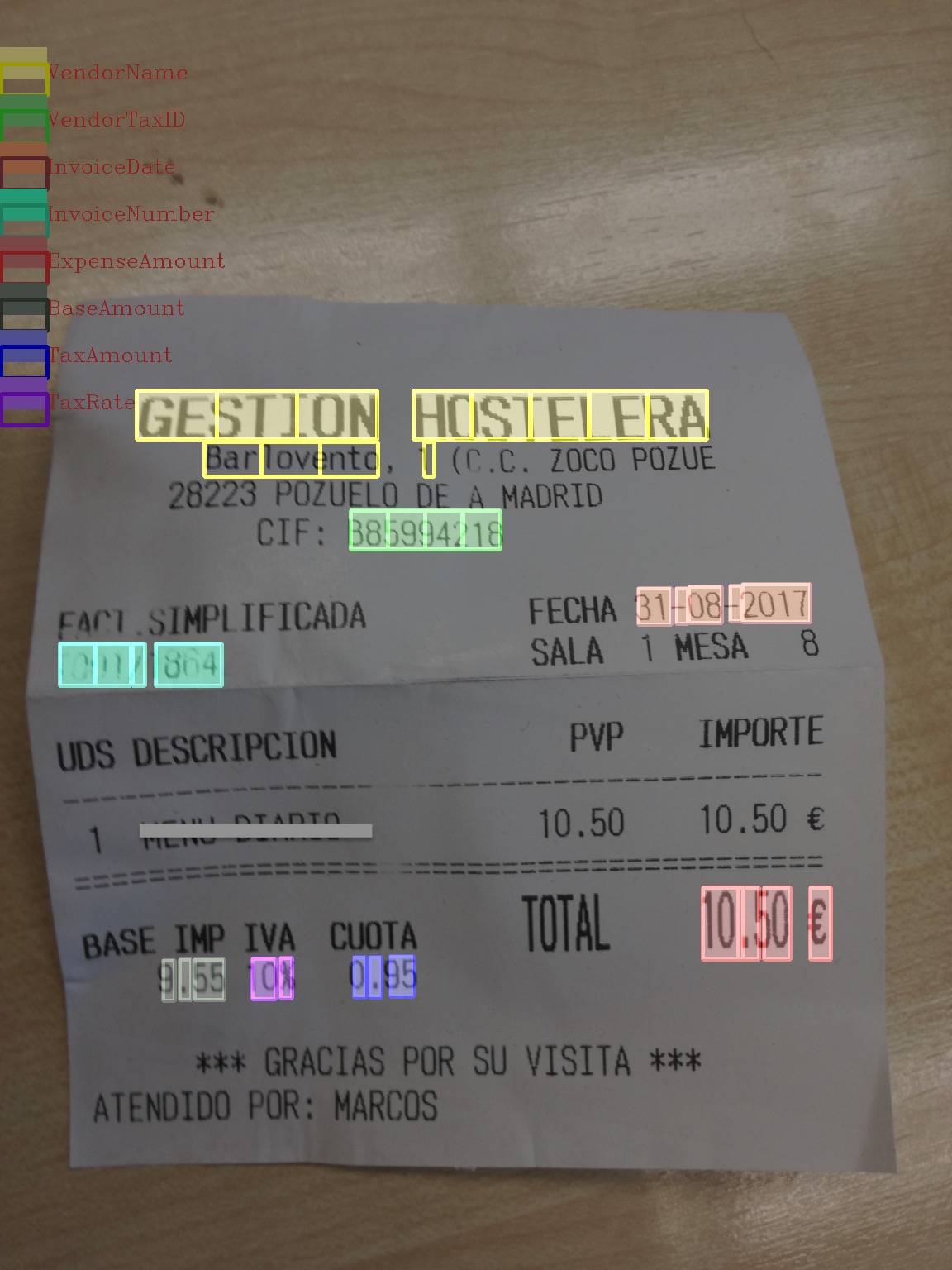}}
\subfloat{\fcolorbox{white}{white}{}}
\subfloat{\includegraphics[width=0.31\linewidth]{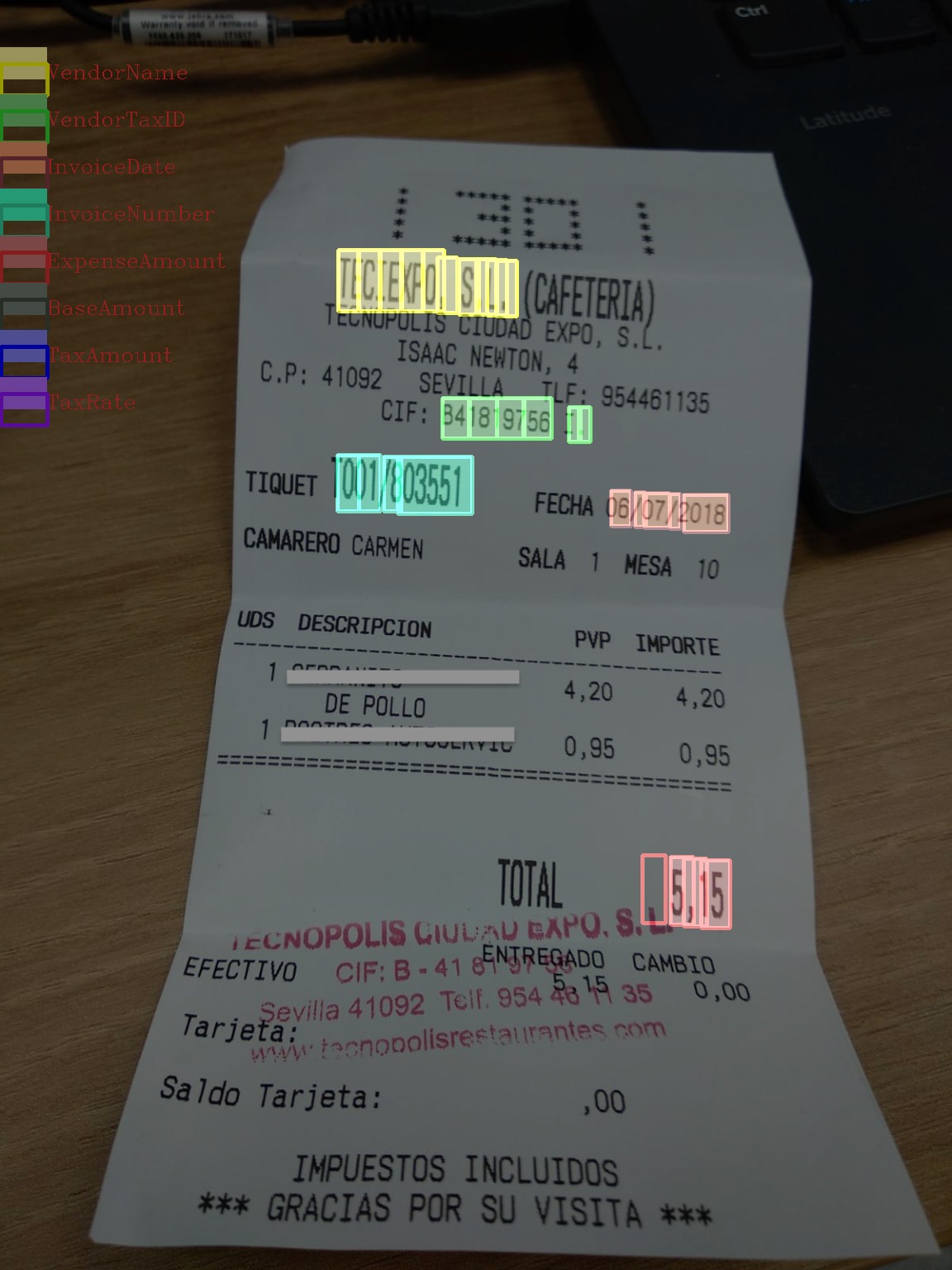}}\\
\subfloat{\includegraphics[width=0.29\linewidth]{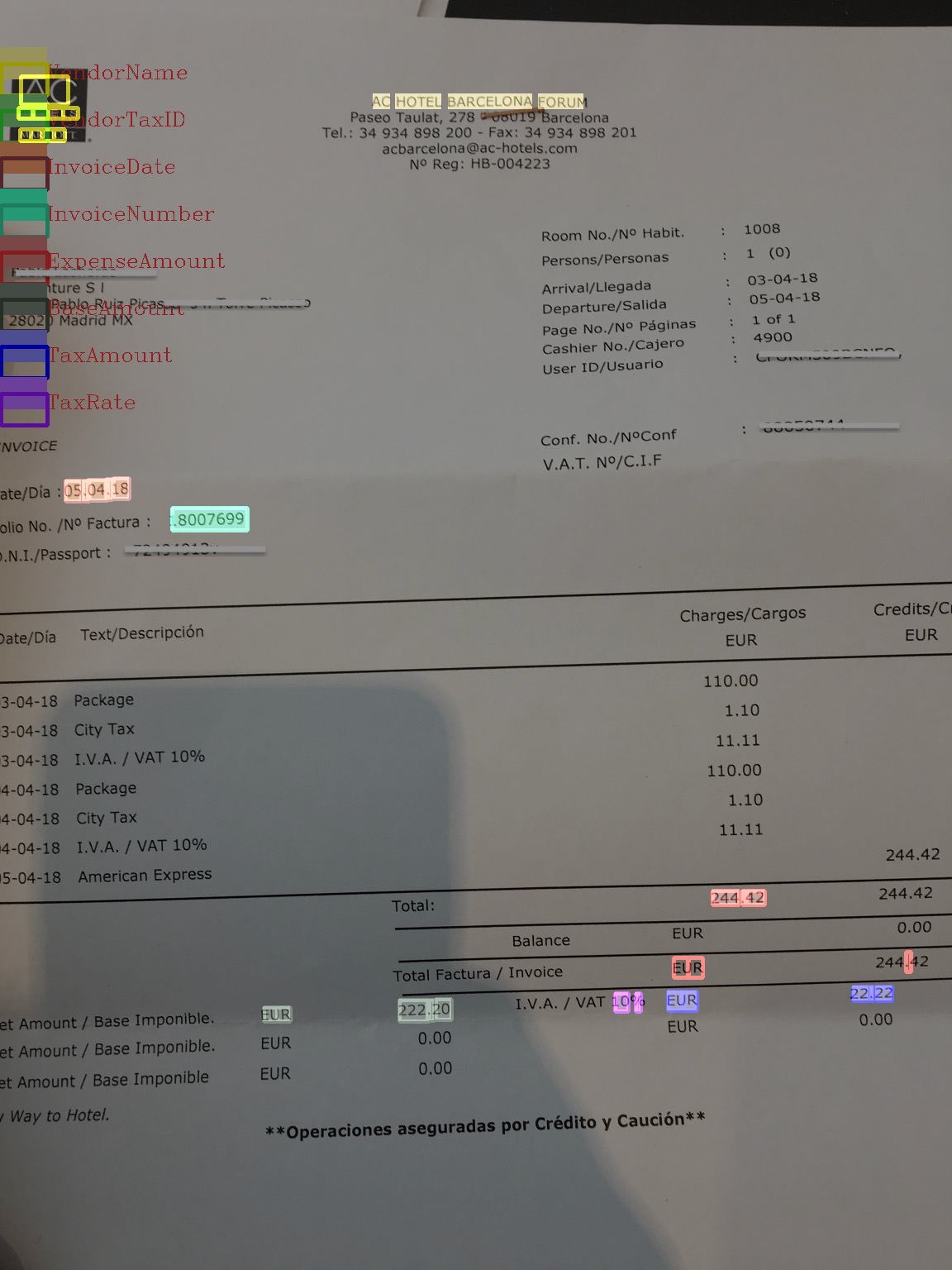}}
\subfloat{\fcolorbox{white}{white}{}}
\subfloat{\includegraphics[width=0.35\linewidth]{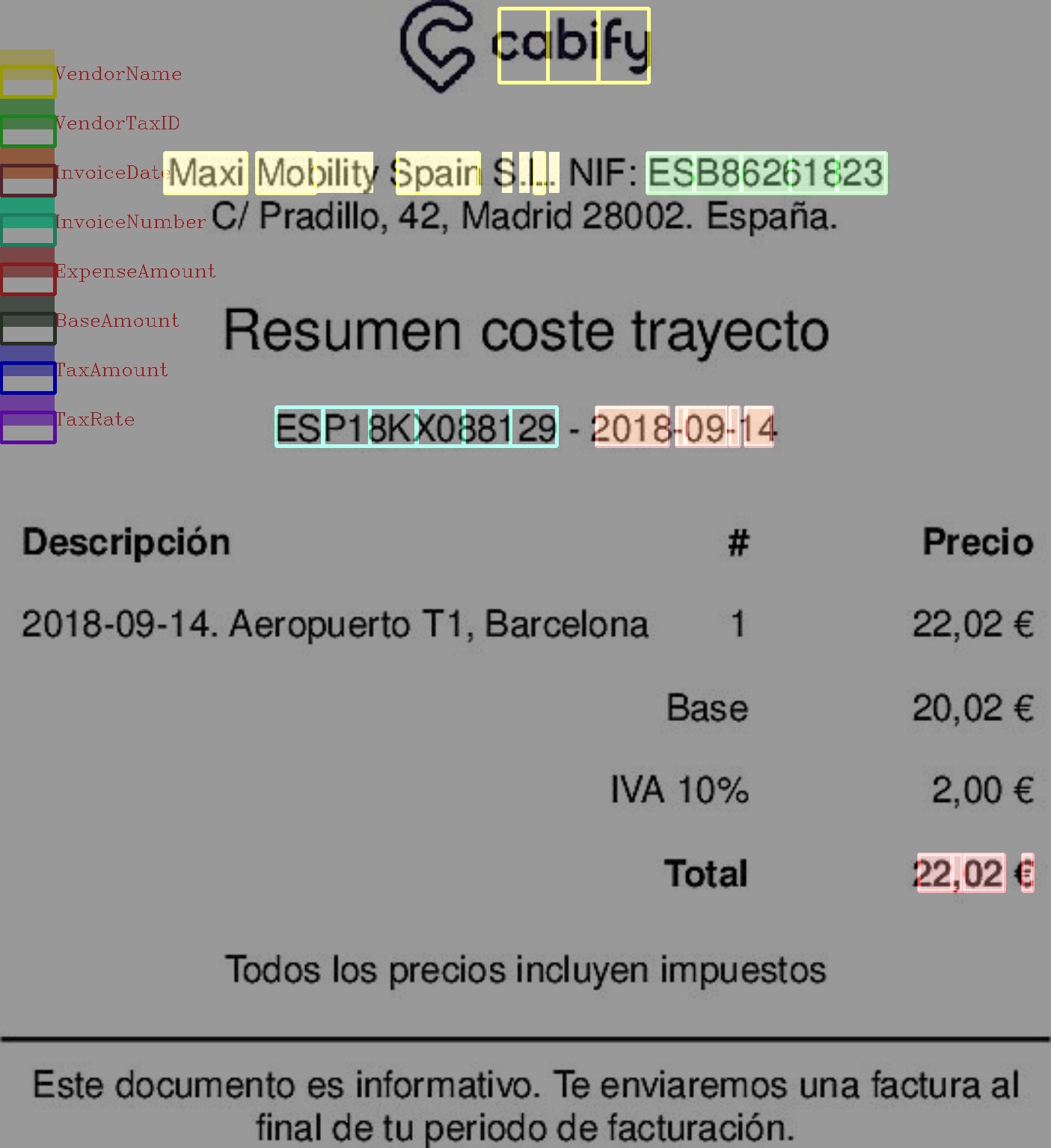}}
\subfloat{\fcolorbox{white}{white}{}}
\subfloat{\includegraphics[width=0.29\linewidth]{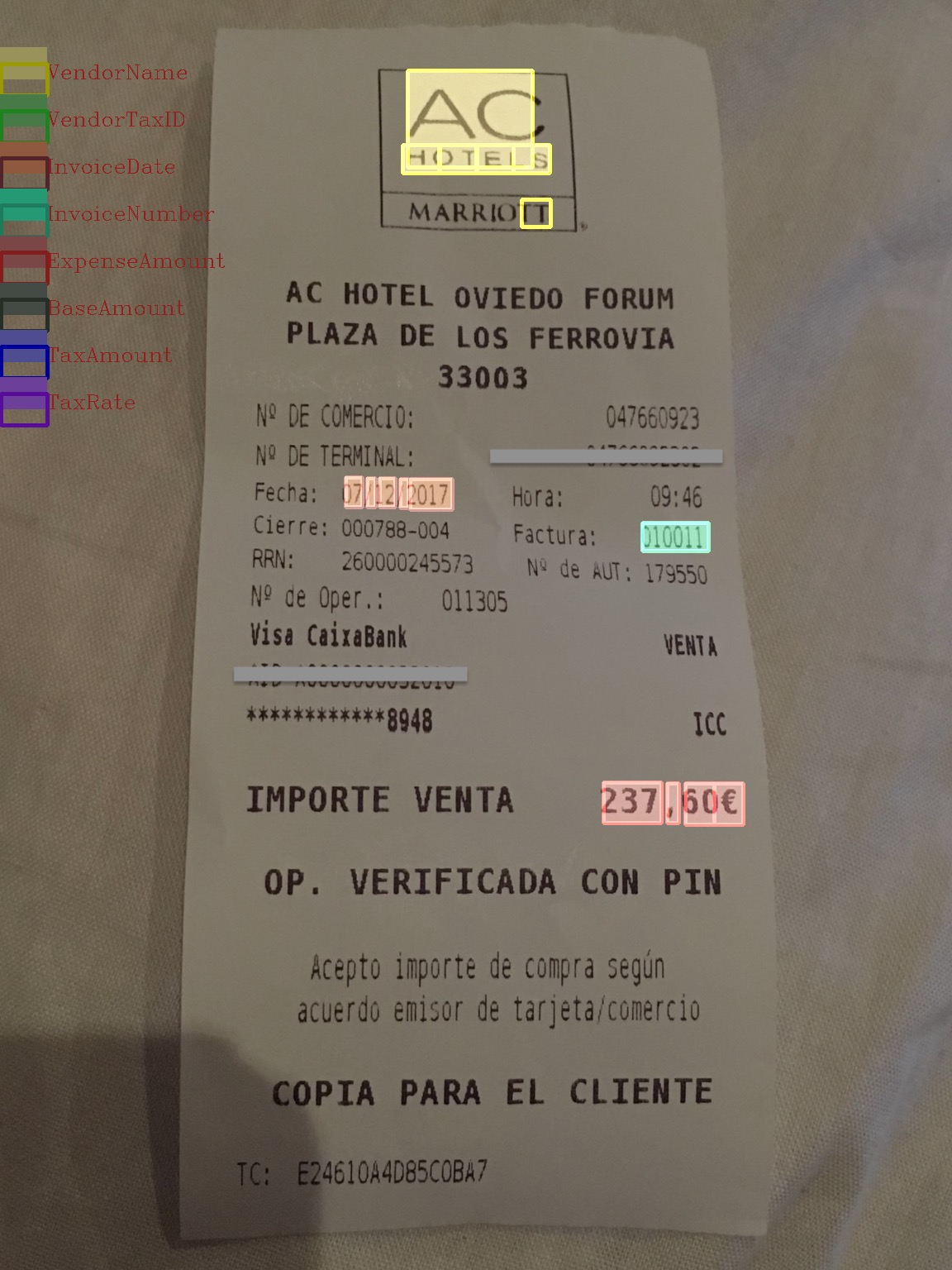}}
\end{center}
   \caption{False positive examples of CUITE prediction results. Color legend in the top-left corner indicates the key information classes. Each color indicates a key information class, where filled rectangles are the ground truths while the boundary-only rectangles are the inference results. The false positives are results with the boundary-only rectangles being not overlapped with the filled rectangles. We mask out certain private information with filled gray rectangles in these figures. (zooming in to check the details)}
\label{fig:falsepositive}
\end{figure}

Typical examples of receipts with low AP but high softAP score are shown in Figure \ref{fig:falsepositive}. Most of the false positive cases occur in the 'VenderName' class, where names tend to vary greatly and leads to difficulty in model inference. However, it is not hard to find that these false positives can be easily avoided by appending a dictionary based post processor to the key information extractor. One rare false positive case, the third receipt in the first row, is a 'L' letter being mis-recognized as '1' by the employed OCR engine. Although the letter shows distinct appearance from the other digits in the scanned receipt image, it is mis-interpreted as one part of the 'BaseAmount' class due to its close spatial location to the digits and it-self being a digit. It also can be seen in the first receipt in the second row that several spatial distant digits were wrongly predicted as in the 'BaseAmount' class. Although these are rare cases in our test set, it still suggests that incorporating image-level information may further boost the inference accuracy inspite of the already involved semantical and spatial features. 

\begin{figure}
\begin{center}
\subfloat{\includegraphics[width=0.28\linewidth]{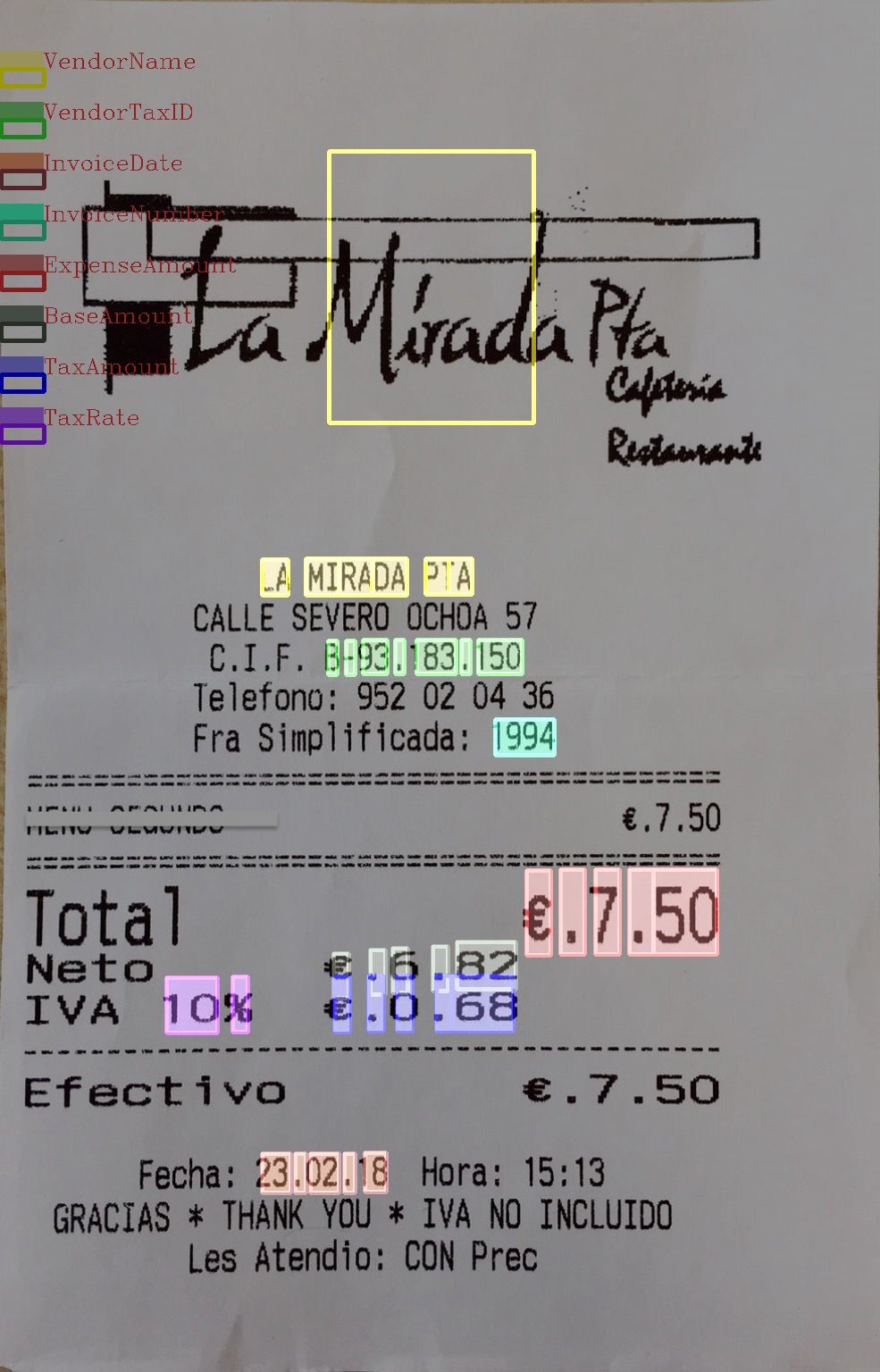}} 
\subfloat{\fcolorbox{white}{white}{}}
\subfloat{\includegraphics[width=0.34\linewidth]{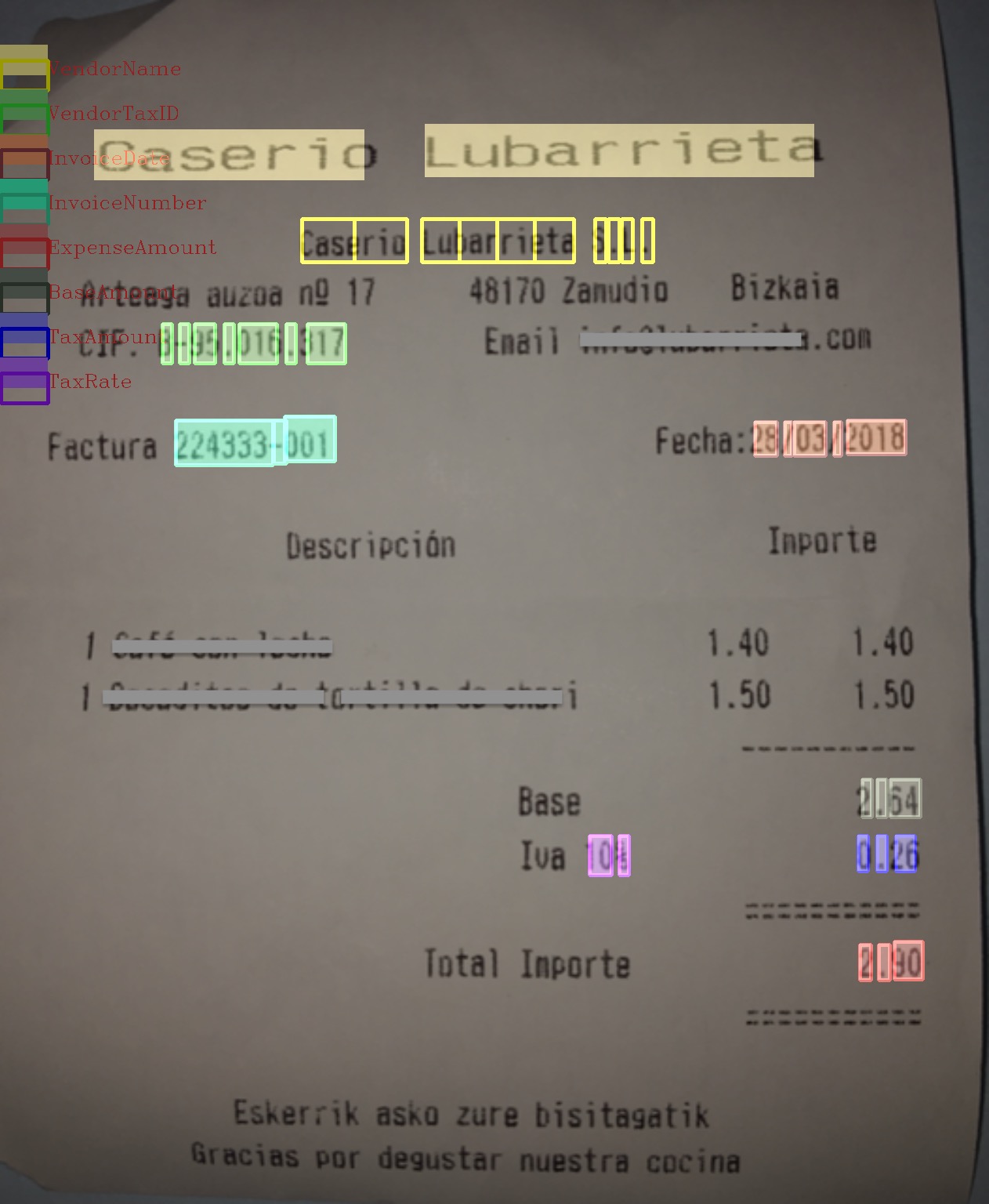}} 
\subfloat{\fcolorbox{white}{white}{}}
\subfloat{\includegraphics[width=0.31\linewidth]{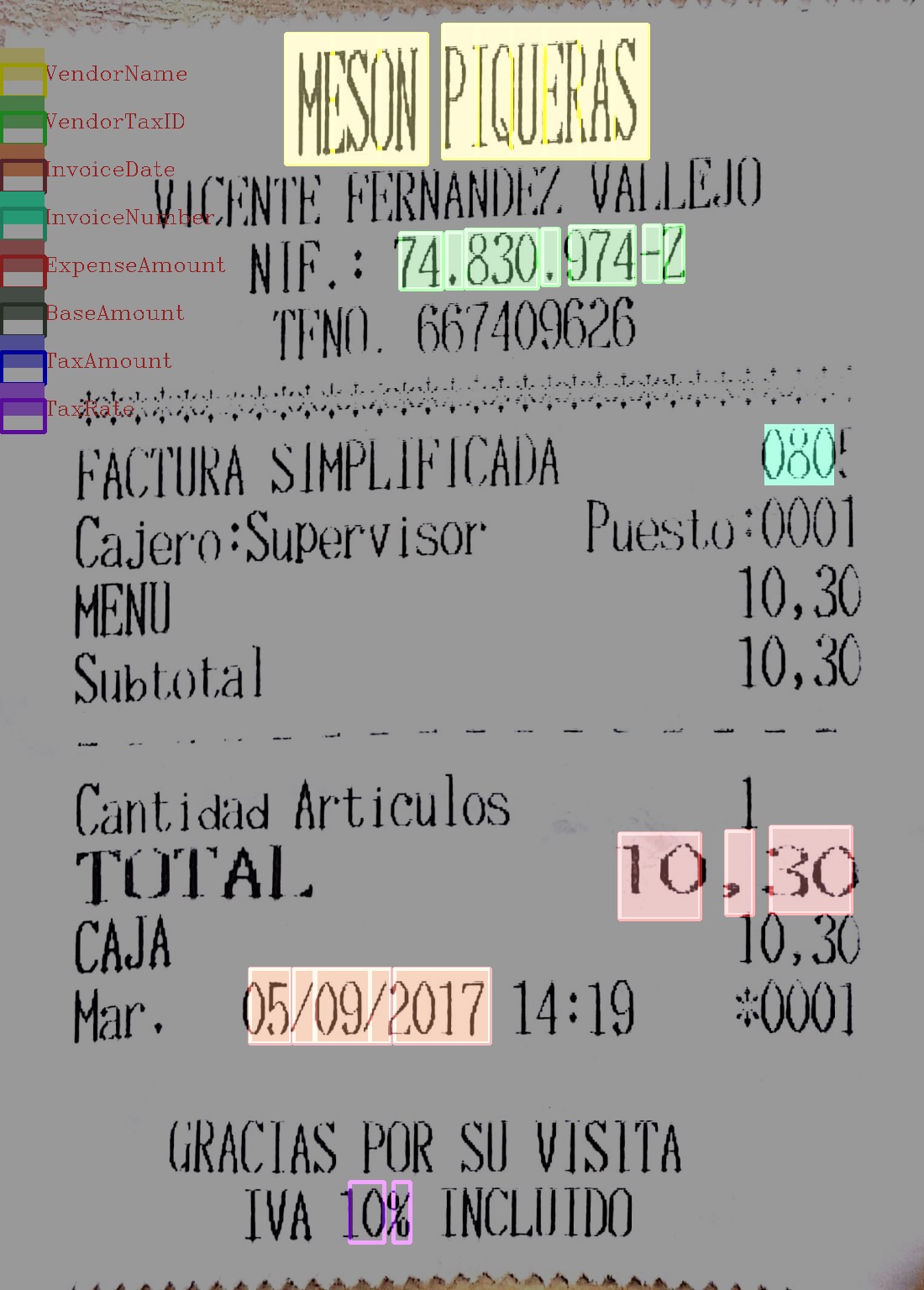}}\\
\subfloat{\includegraphics[width=0.31\linewidth]{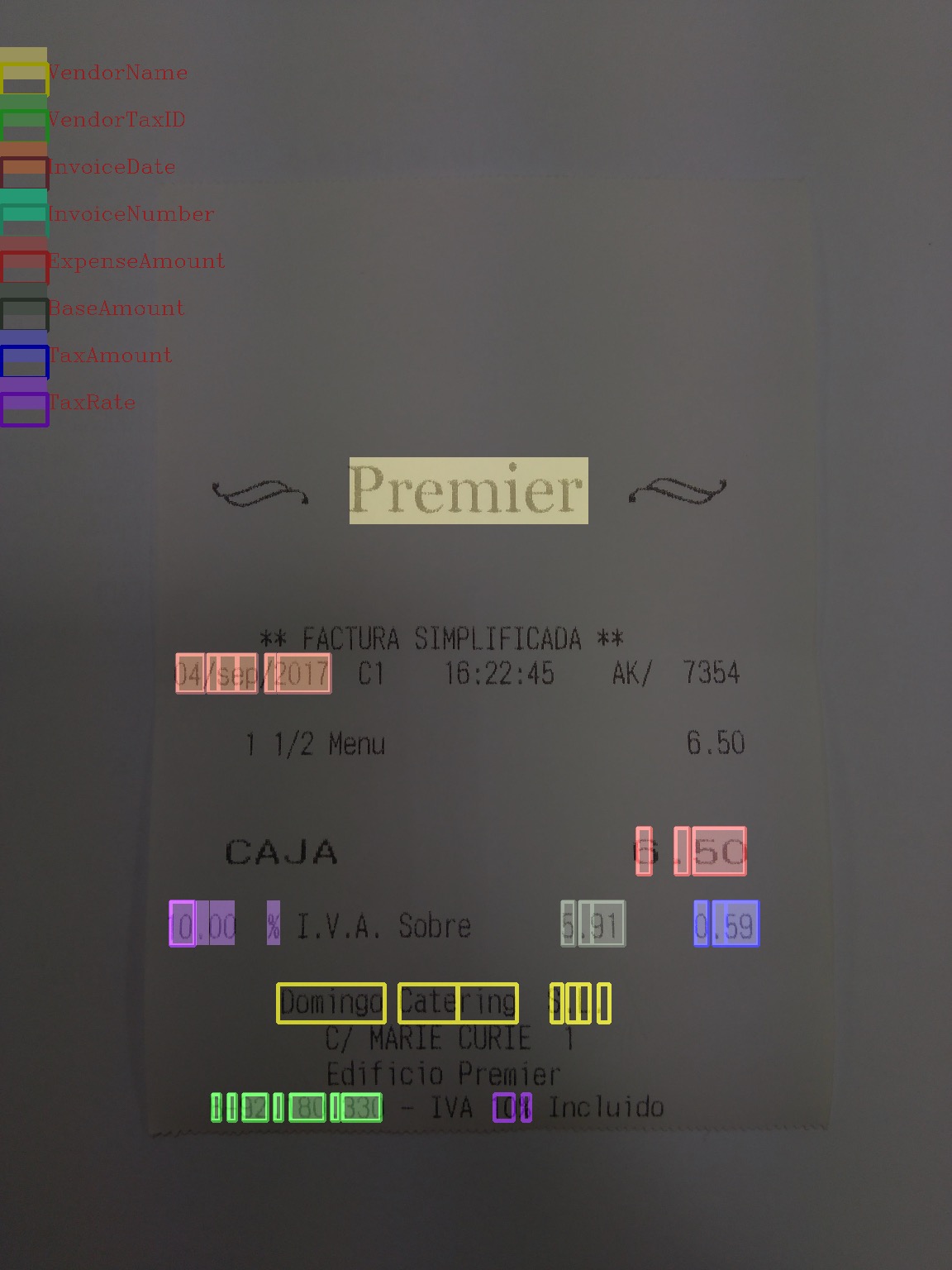}} 
\subfloat{\fcolorbox{white}{white}{}}
\subfloat{\includegraphics[width=0.31\linewidth]{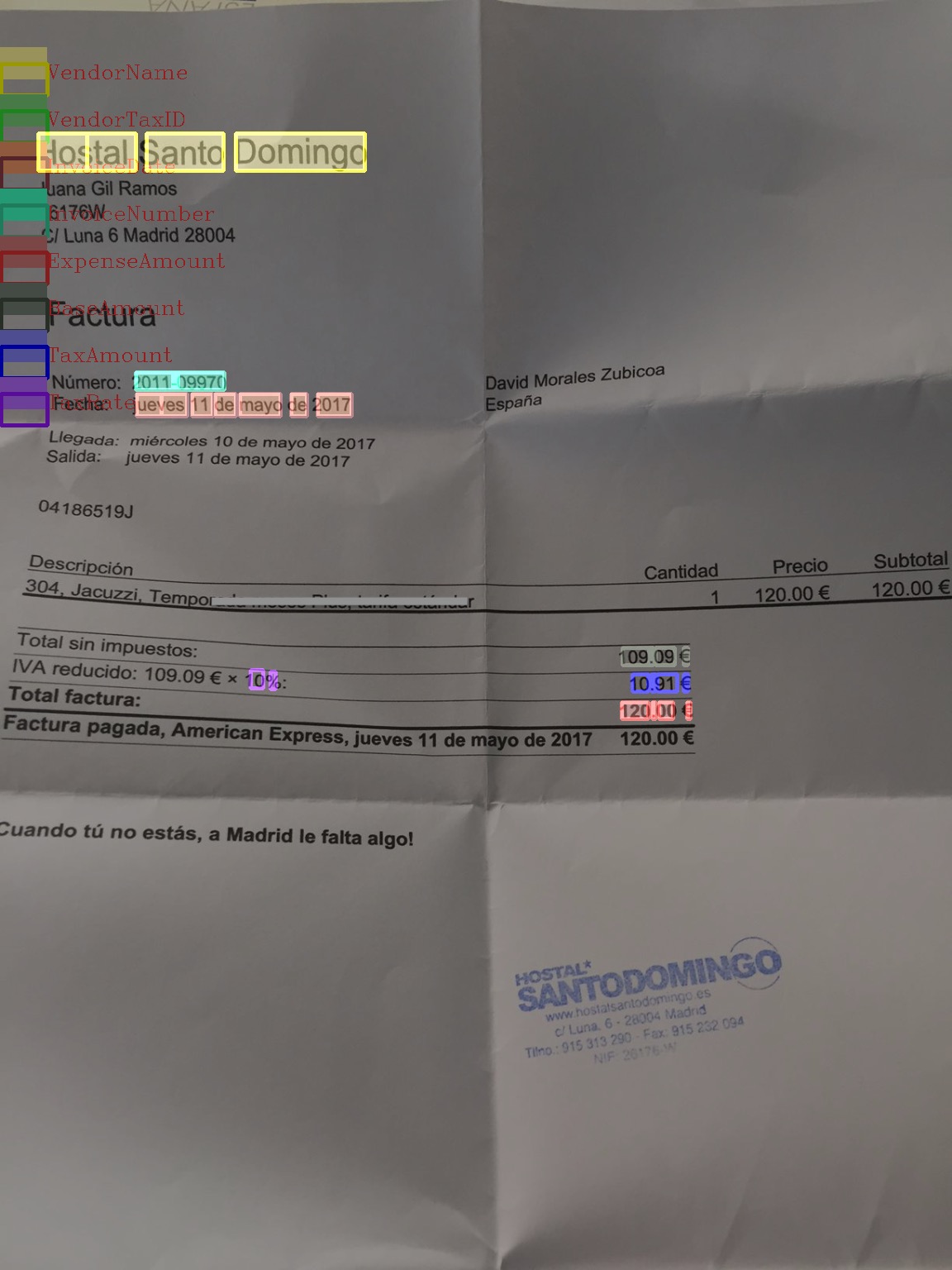}}
\subfloat{\fcolorbox{white}{white}{}}
\subfloat{\includegraphics[width=0.31\linewidth]{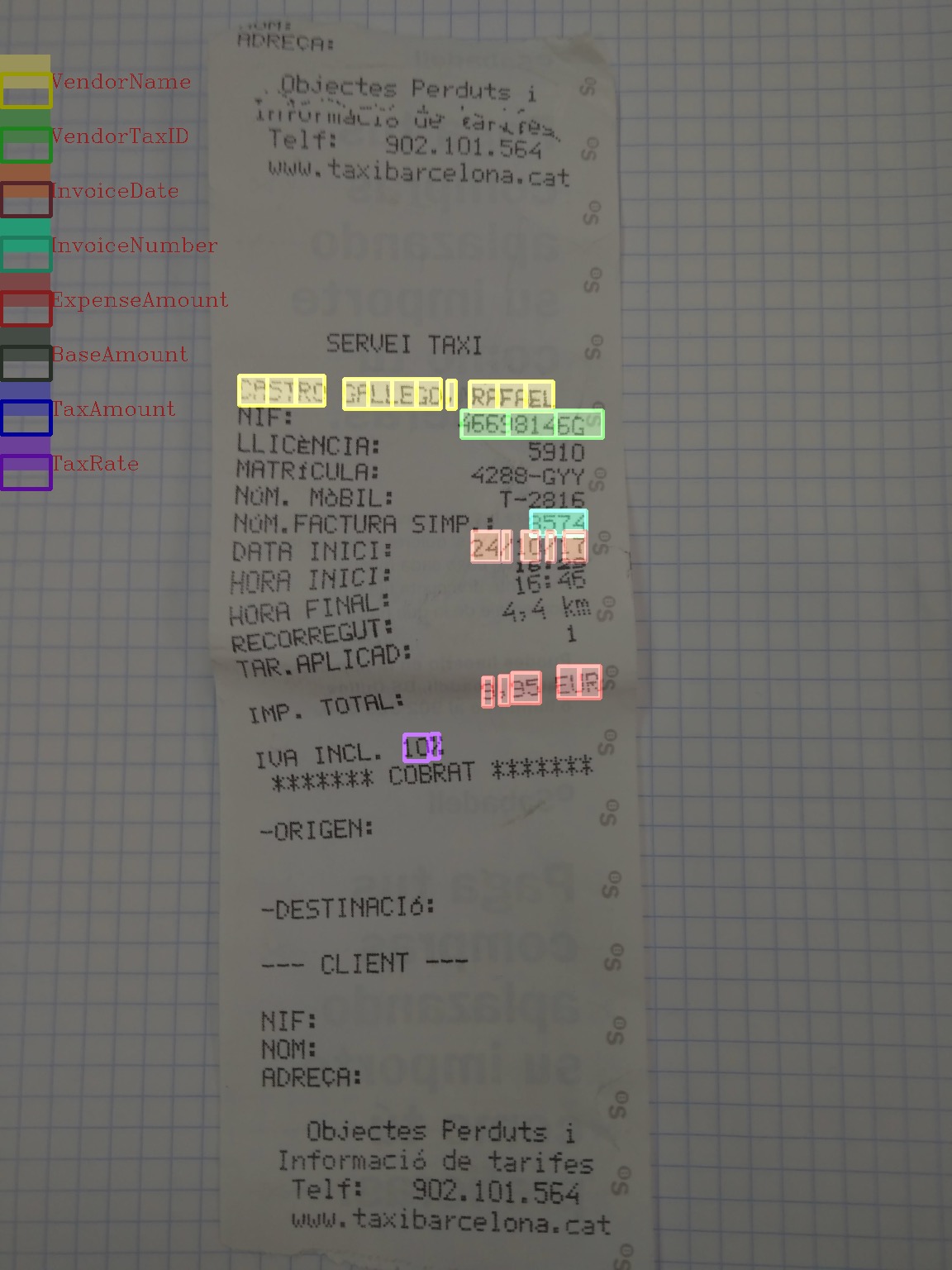}}\\
\subfloat{\includegraphics[width=0.36\linewidth]{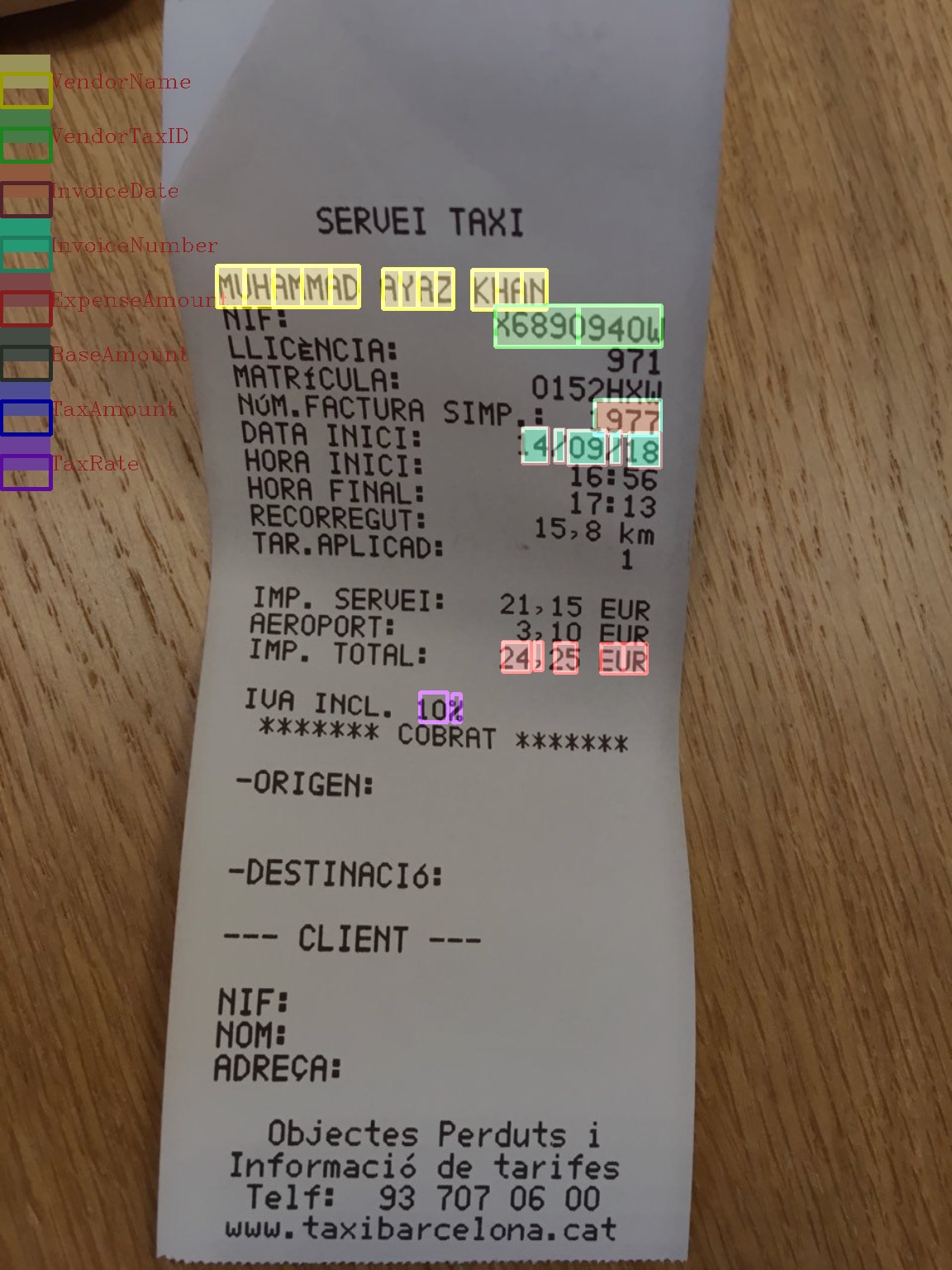}} 
\subfloat{\fcolorbox{white}{white}{}}
\subfloat{\includegraphics[width=0.22\linewidth]{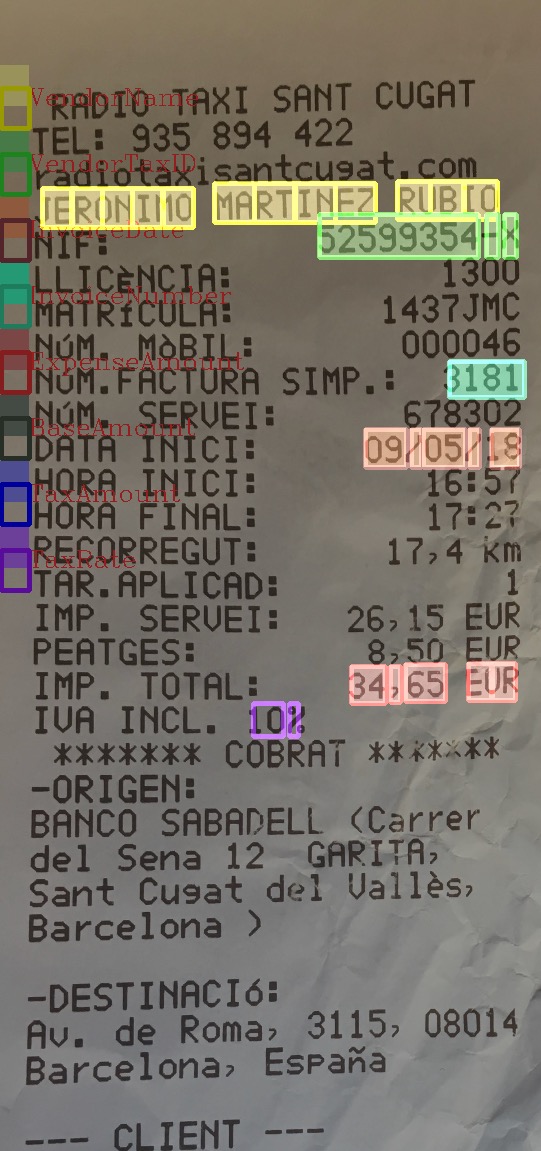}}
\subfloat{\fcolorbox{white}{white}{}}
\subfloat{\includegraphics[width=0.36\linewidth]{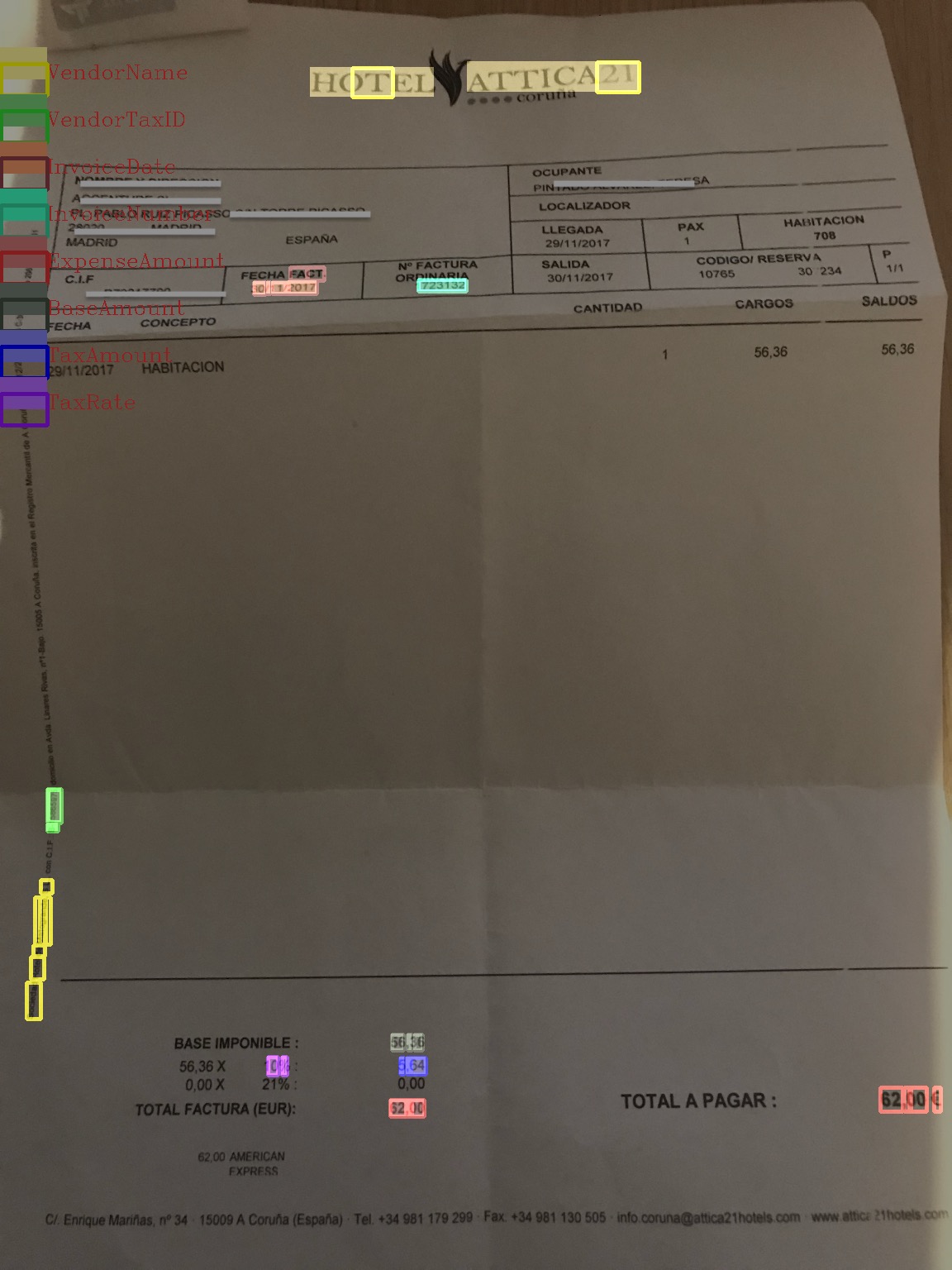}}\\
\end{center}
   \caption{False labelling examples of CUITE prediction results, where the error is actually caused by the wrong labelling of human labeler. Color legend in the top-left corner indicates the key information classes. Each color indicates a key information class, where filled rectangles are the ground truths while the boundary-only rectangles are the inference results. We mask out certain private information with filled gray rectangles in these figures. (zooming in to check the details)}
\label{fig:falselabel}
\end{figure}

Furthermore, we find that some wrong cases are actually correct since the ground truths were wrongly labelled by the human labelers. As illustrated in Figure \ref{fig:falselabel}, the 'VendorName' appears twice in the scanned image but only being labelled once in both the first and second receipt in the first row, while the model correctly interprets both occurance as the 'VendorName' in the first receipt and correctly interprets the second occurance in the second receipt. Furthermore, 'TaxRate' is neglected to be labelled in the third receipt in the first row and 'VendorName' is wrongly labelled in the first receipt in the second row. The third receipt in the second row and the first and second receipt in the third row are all neglected with labelling 'TaxRate', and the third receipt in the third row is neglected with labelling 'BaseAmount' while the trained model correctly inferred the right class. It is not hard to find that the trained model produces even better results than human labeler on these receipts, which further proves the effectiveness of the proposed method.

\begin{table*}
	\caption{Performance evaluation of CUTIE on ME with different embedding size.}
\begin{center}
\begin{tabular}{l | c | c | c | c | c | c | c | c | c | c}
	Embedding Size & 1 & 2 & 4 & 8 & 16 & 32 & 64 & 128 & 256 & 512 \\
	\hline
	\#Params & 10.6M & 10.6M & 10.7M & 10.7M & 10.9M & 11.3M & 12.1M & 13.6M & 16.6M & 22.7M \\
	AP & 79.1 & 82.8 & 83.1 & 82.8 & 82.9 & 83.2 & 83.8 & 84.3 & \textbf{84.6} & 84.4 \\
	softAP & 87.3 & 90.3 & 90.4 & 92.0 & 90.6 & 90.7 & 92.3 & \textbf{92.4} & 91.9 & 91.9 \\
\end{tabular}
\end{center}
	\label{tab:embedding}
\end{table*}

\begin{table*}
	\caption{Performance evaluation of CUTIE-B on ME with different number of training samples.}
\begin{center}
\begin{tabular}{l | c | c | c | c | c | c | c | c | c}
	Percentage(\%) & 3 & 12 & 21 & 30 & 39 & 48 & 57 & 66 & 75 \\
	\hline
	AP & 56.2 & 76.4 & 79.6 & 80.8 & 82.6 & 81.4 & 83.4 & \textbf{85.0} & 84.3 \\
	softAP & 76.0 & 86.5 & 88.7 & 90.3 & 90.5 & 90.7 & 90.7 & 91.1 & \textbf{91.5} \\
\end{tabular}
\end{center}
	\label{tab:samples}
\end{table*}

We also evaluate the CUTIE-B model on the ICDAR 2019 SROIE trainval set and list the results in Table \ref{tab:sroie}. The result in class level indicates that the inferenced result is perfectly the same with the ground truth, whereas the result in soft class level indicates that the inferenced result incorporates the ground truth but also included some unrelated text tokens. The result in token level indicates the token classification accuracy of each class in spite of the 'DontCare' class.
\begin{table*}
	\caption{Performance evaluation of CUTIE-B on ICDAR 2019 SROIE task 3.}
\begin{center}
\begin{tabular}{l | c | c | c}
	 & class level & class level (soft) & token level \\
	\hline
	CUTIE-B & 86.7\% & 92.7\% & 94.2\% \\
\end{tabular}
\end{center}
	\label{tab:sroie}
\end{table*}

\subsection{Ablation Studies}
Although we have demonstrated extremely strong empirical results, the results presented are achieved by combination of each aspect of the CUTIE framework. In this section, we perform ablation studies over a number of facets of CUTIE in order to have a better understanding of their relative importance. CUTIE-B is employed as the default model with grid augmentation, embedding size of $128$. In this section, we use all data of ME, where $75\%$ of them are used to train the model and the rest $25\%$ for test.

\subsubsection{Effect of Grid Augmentation on Understanding Spatial Information}
One of our core claims is that the high performance CUTIE is achieved by jointly analysis of the semantical and spatial information with the proposed framework. The highly effective spatial analysis ability is enabled by the CUTIE framework in contrast with the previous NER based methods. The grid augmentation process further enhances this ability. To provide more evidence to this claim, we evaluate CUTIE with or without the grid augmentation process to test the model performance in terms of spatial diversity. Results are presented in Table \ref{tab:augmentation}. We can see that adding the grid augment process in CUTIE significantly improves the performance. These results demonstrate that CUTIE can greatly benefit from enhancing data diversity in spatial distribution. For that reason, further analysis about grid augmentation techniques may enhance CUTIE's performance, \eg, randomly moving certain texts upward, downward, leftward, or rightward by several pixels during the grid positional mapping process. 
\begin{table}
	\caption{Performance evaluation of CUTIE on ME with or without the grid augmentation process.}
\begin{center}
\begin{tabular}{l | c | c}
	 & w/o augmentation & w augmentation \\
	\hline
	AP & 83.3 & 84.3 \\
	softAP & 93.7 & 92.4 \\
\end{tabular}
\end{center}
	\label{tab:augmentation}
\end{table}

\subsubsection{Impact of Semantical Information Capacity}
\label{parameters}
Next, we evaluate the impact of semantical information of CUTIE by comparing evaluation with different word embedding sizes. As reported in Table \ref{tab:embedding}, CUTIE performs in a bell shape curve as the embedding size increases. The best performance is achieved by CUTIE-B with $85.0\%$ in AP and $92.9\%$ in softAP using embedding size of $64$. One interesting finding is that the CUTIE model achieves good performance even with limited semantical information capacity. We infer the reason is that, for the SROIE problem with 9 key information classes, there is a small amount of key token providing the majority contribution to the model inference and the model can achieve good performance by paying special attention to these key tokens. It is also indicated in the result that too large embedding size also decreases the model performance. 

\subsubsection{Impact of Number of Training Samples}
To evaluate the impact of using different number of training samples on model performance, we train CUTIE with $3\%$, $12\%$, $21\%$, $30\%$, $39\%$, $48\%$, $57\%$, $66\%$, and $75\%$ of our dataset and report results in Table \ref{tab:samples}. It is not hard to find from the result that more training data leads to better performance either in terms of AP or softAP. CUTIE-B achieves the highest AP $85.0\%$ with $66\%$ of dataset as training samples and the highest softAP $91.5\%$ with $75\%$ of dataset as training samples. It is worth noting that CUTIE-B is already capable of achieving $79.6\%$ AP and $88.7\%$ softAP with only $21\%$ of dataset as training samples, which further proves the efficiency of the proposed method being capable of achieving good results with a limited amount of training data.

\subsubsection{Other Experiments}
\begin{figure}
\begin{center}
\subfloat{\includegraphics[width=0.31\linewidth]{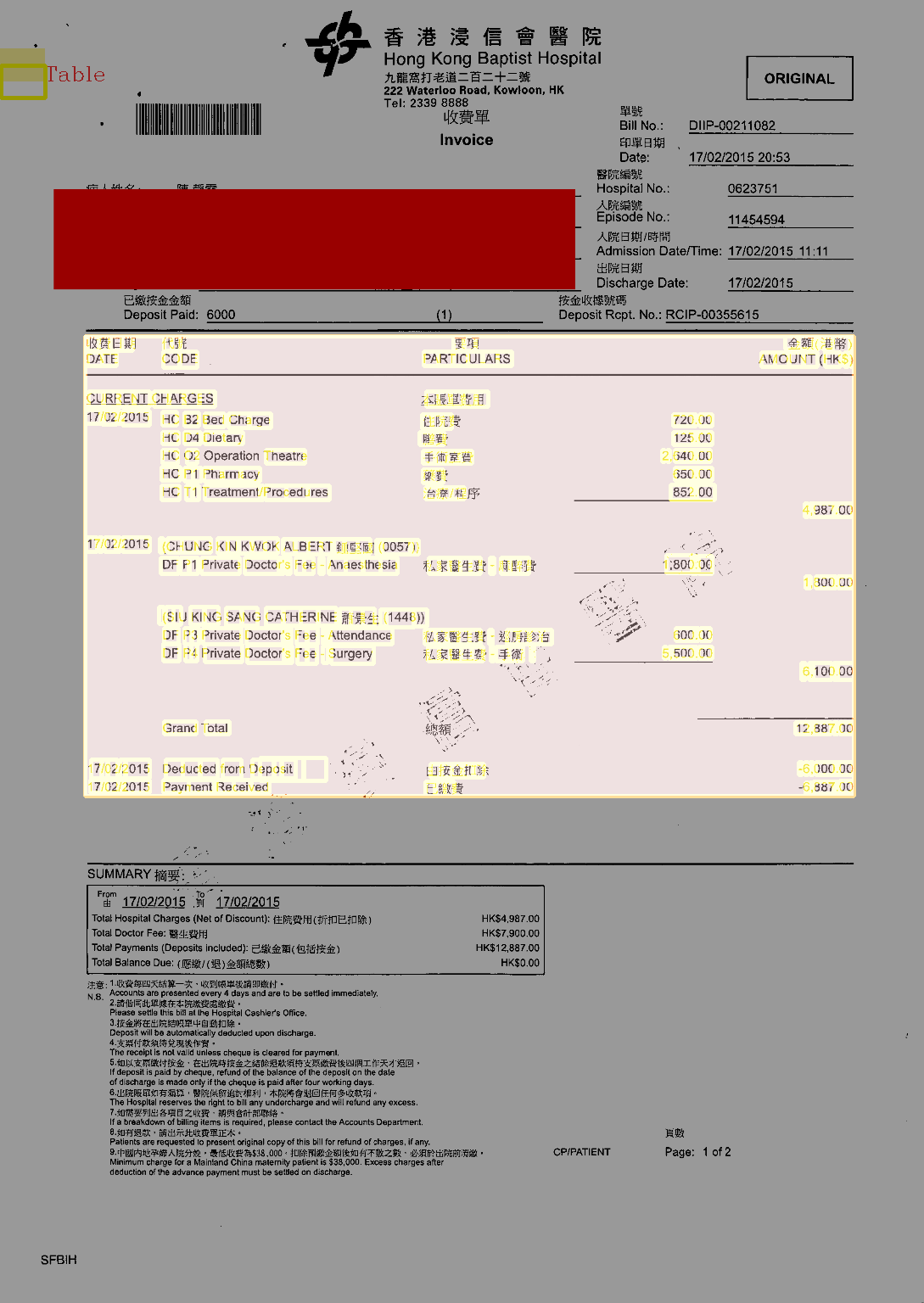}} 
\subfloat{\fcolorbox{white}{white}{}}
\subfloat{\includegraphics[width=0.31\linewidth]{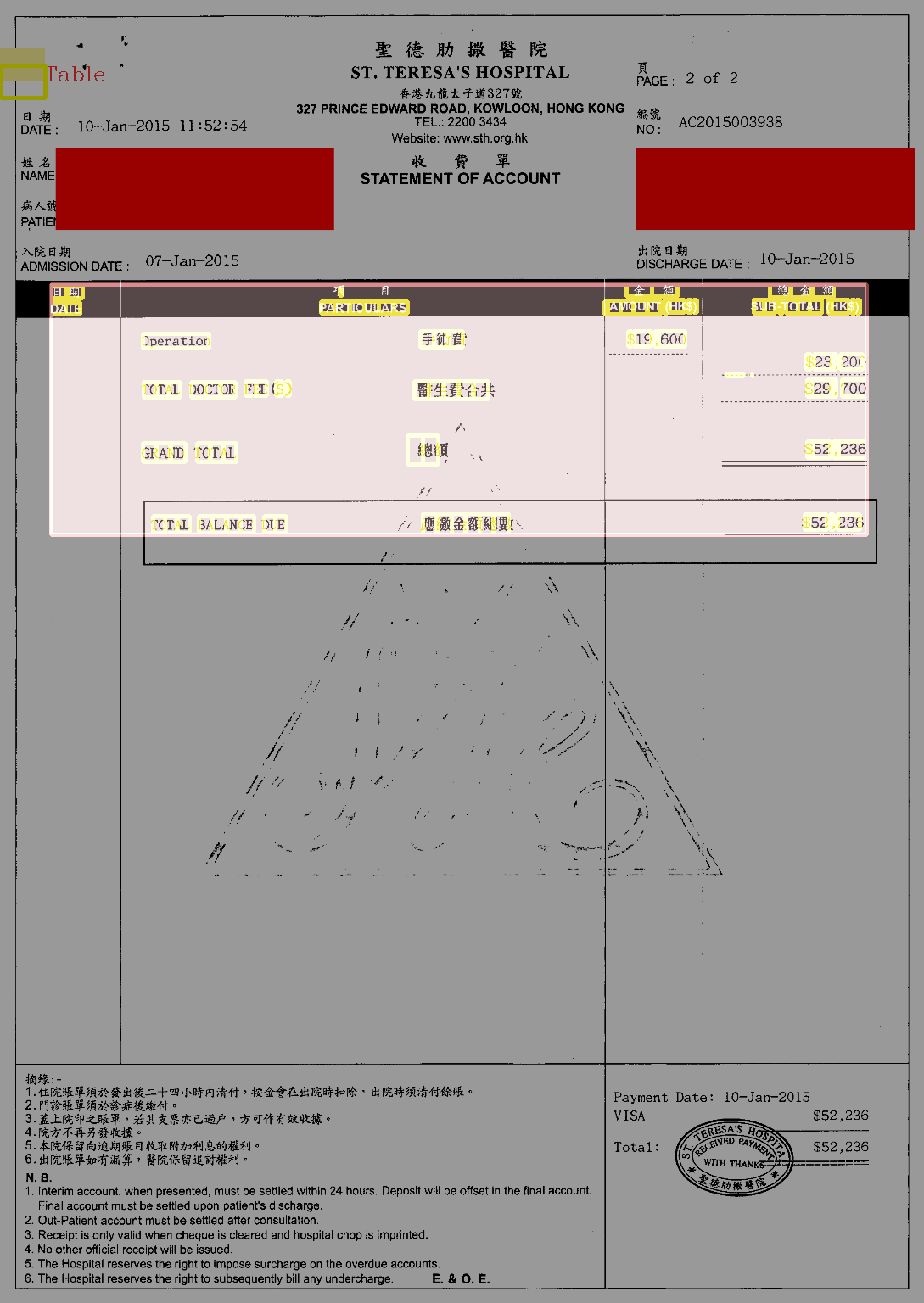}} 
\subfloat{\fcolorbox{white}{white}{}}
\subfloat{\includegraphics[width=0.31\linewidth]{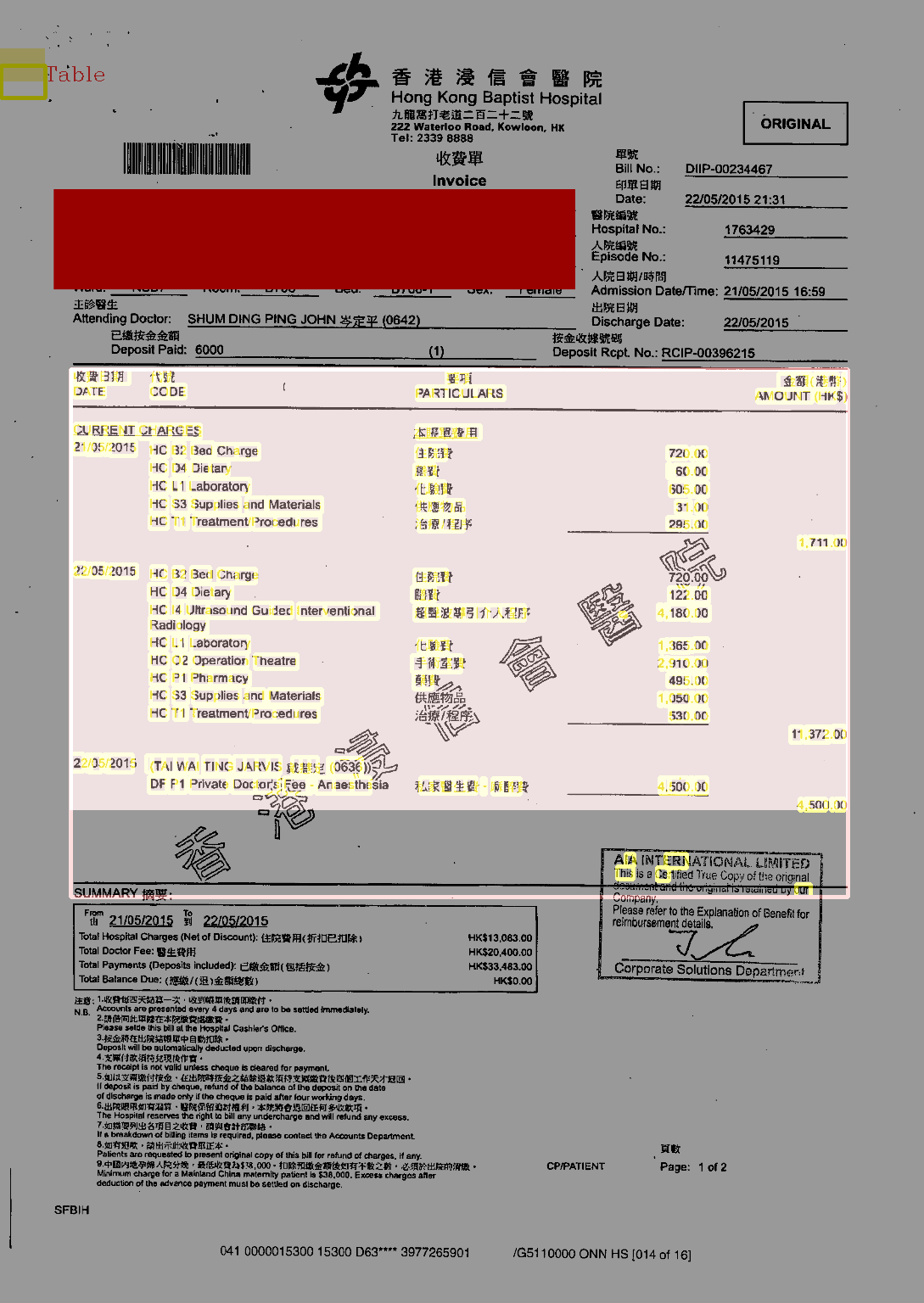}} 
\end{center}
   \caption{Locating the table region with CUTIE-B. Table tokens are indicated with yellow bounding boxes and table regions are indicated with pink boxes, where filled rectangles are the ground truths while the boundary-only rectangles are the inference results. We mask out certain private information with filled red rectangles in these figures. (zooming in to check the details)}
\label{fig:table}
\end{figure}

\begin{figure}
\begin{center}
\subfloat{\includegraphics[width=0.31\linewidth]{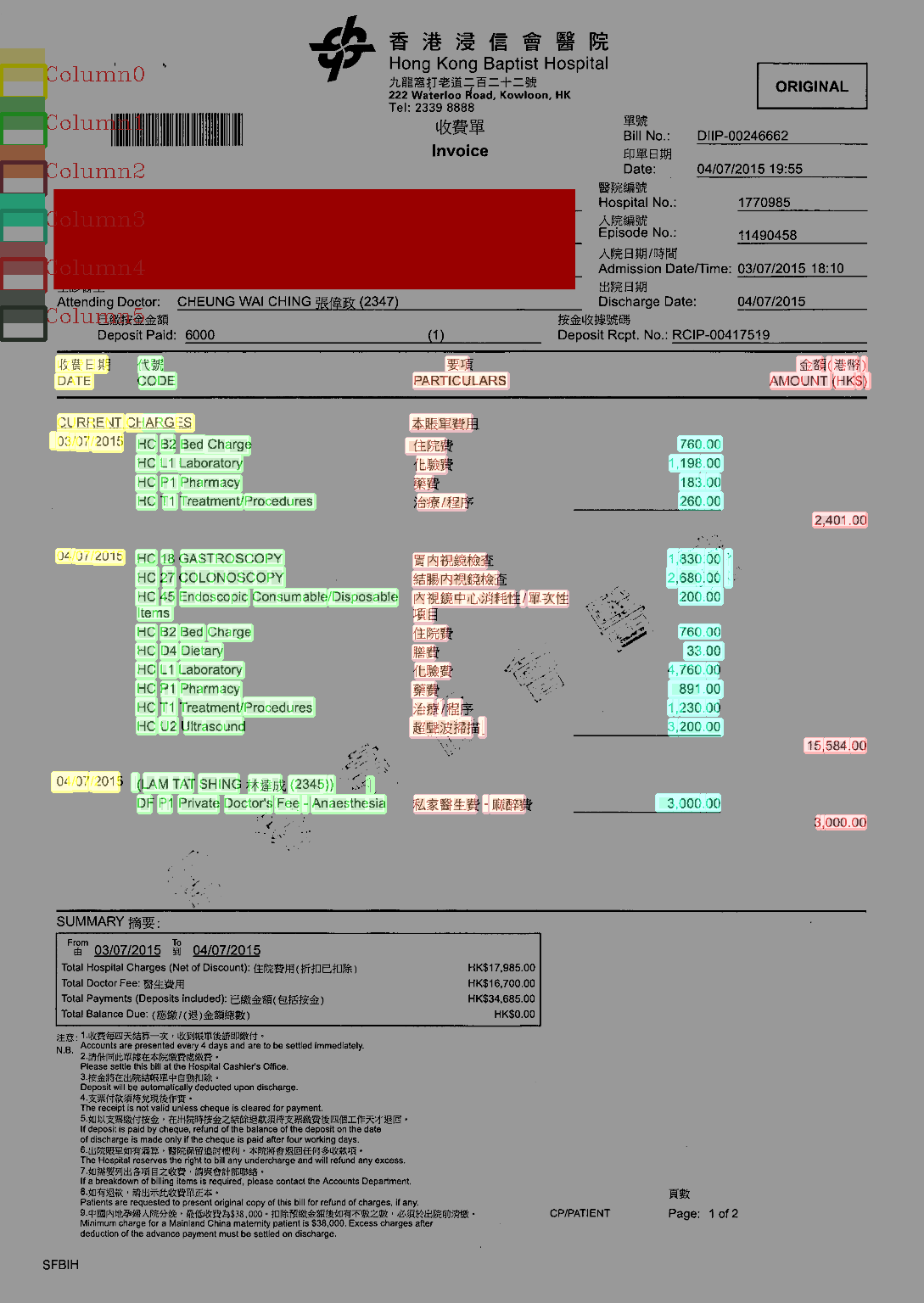}} 
\subfloat{\fcolorbox{white}{white}{}}
\subfloat{\includegraphics[width=0.31\linewidth]{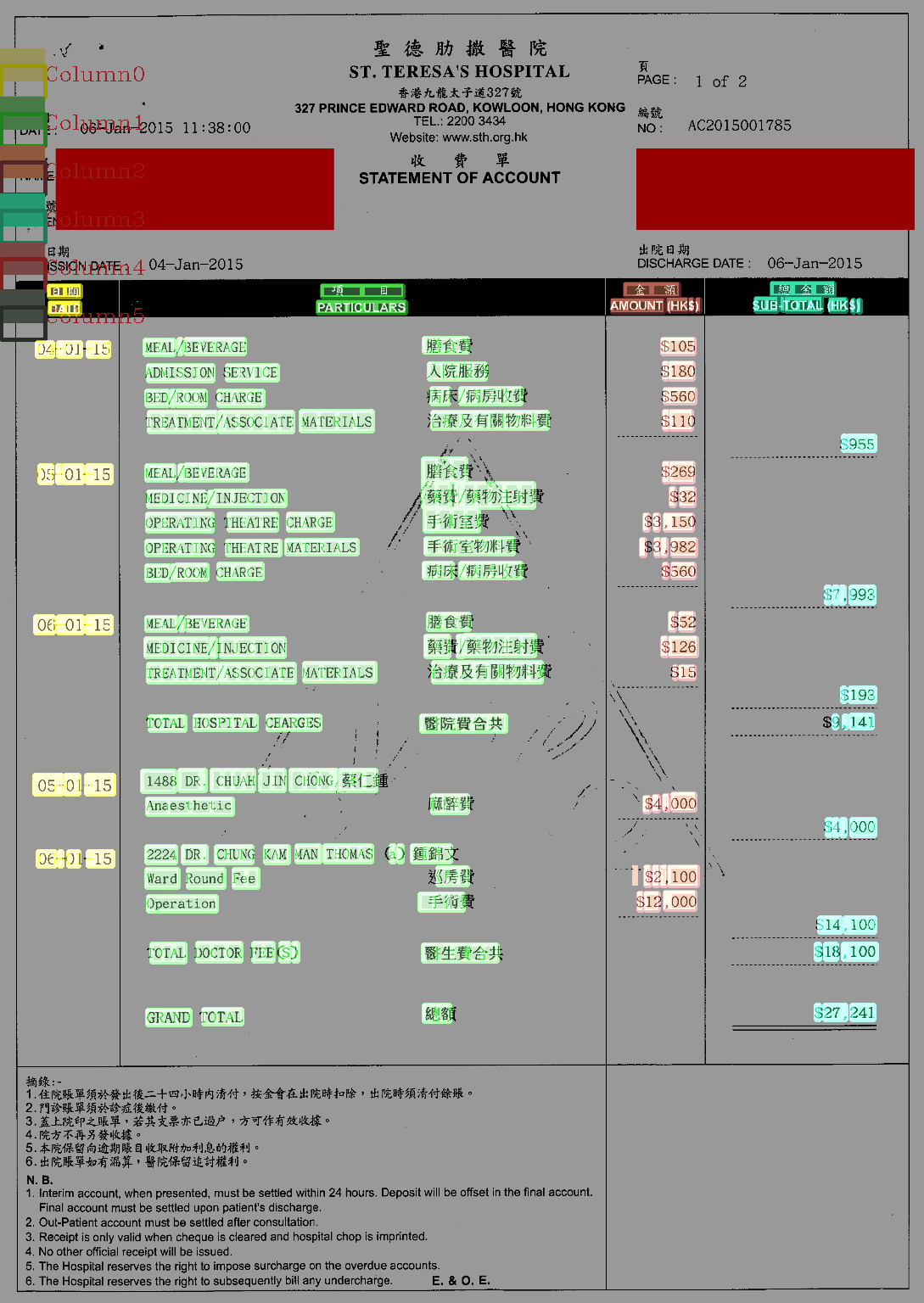}} 
\subfloat{\fcolorbox{white}{white}{}}
\subfloat{\includegraphics[width=0.31\linewidth]{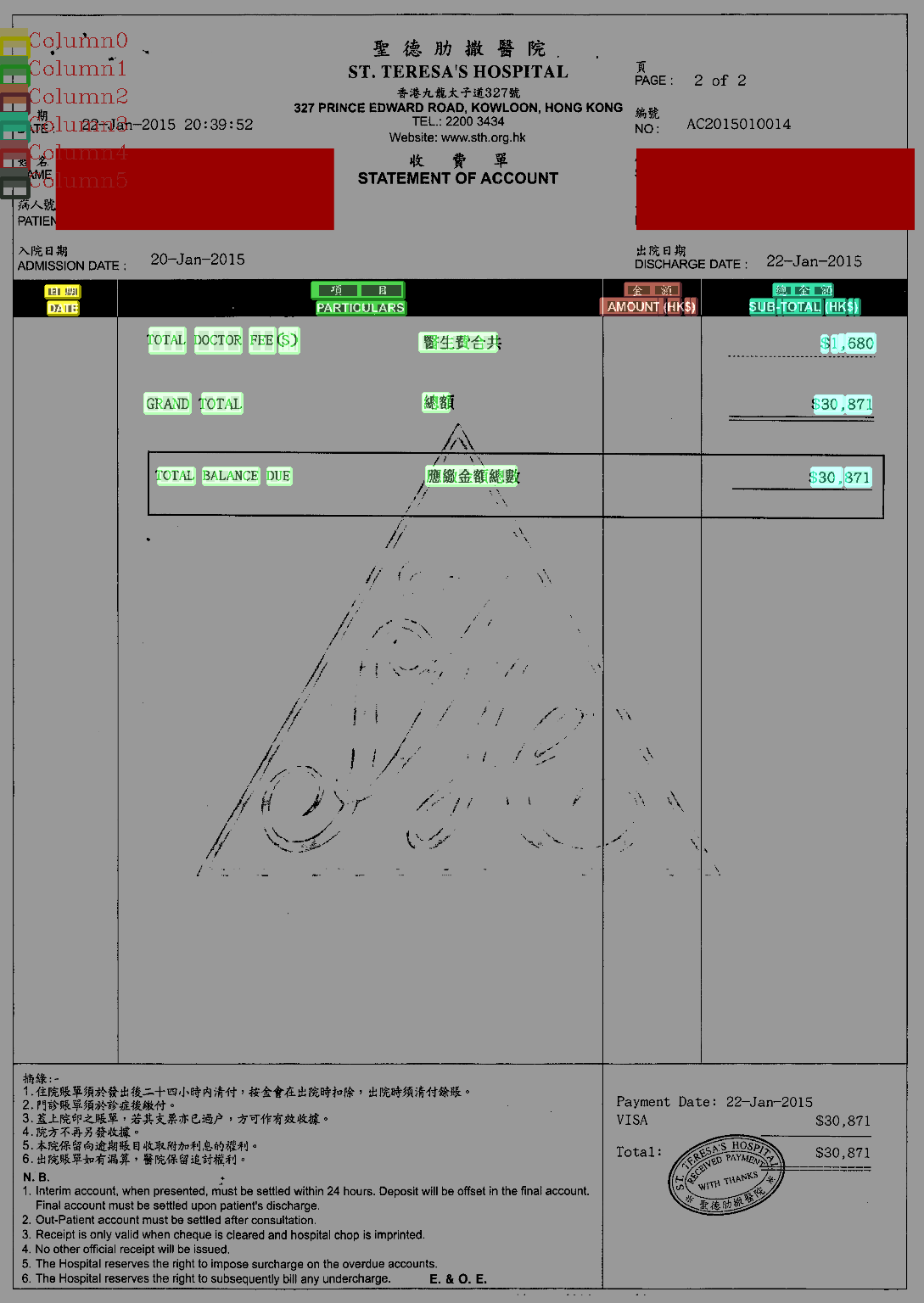}} 
\end{center}
   \caption{Identifying column tokens with CUTIE-B. Color legend in the top-left corner indicates the column id, where filled rectangles are the ground truths while the boundary-only rectangles are the inference results. We mask out certain private information with filled red rectangles in these figures. (zooming in to check the details)}
\label{fig:column}
\end{figure}

Here, we introduce another dataset with $613$ image samples where a text token is labelled with whether it is contained in a table region or not and which table column it belongs to. To test the performance of the proposed method on different tasks, two CUTIE-B models are trained, one for locating the table region in document images and another one for identifying the specific column that the token belongs to. $463$ samples are used for training and $150$ samples for testing. For the table region locating test, experimental results shows that $86\%$ tables are perfectly located, where most of the error cases are due to incorrectly incorporting tokens that not belong to the true table region, as illustrated in Figure \ref{fig:table}. For the column identifying test, $94.8\%$ tokens are perfectly inferenced, as illustrated in Figure \ref{fig:column}.

\section{Discussion}
Automatically extracting interested words / information from the scanned document images is of great interest to various services and applications. This paper proposes CUTIE to tackle this problem without requirement of any pre-training or post processing. Experimental results verify the effectiveness of the proposed method. In contrast to the previous methods, the proposed method is easy to train and requires much smaller amount of training data while achieving state of the art performance with processing speed up to 50 samples per second. The performance gain is mainly achieved by exploring three key factors: the spatial relationship among texts, the semantic information of texts, and the grid positional mapping mechanism. One interesting finding is that the trained CUTIE model correctly predicts certain key information that is neglected to be labelled by the human labeler, which further proves the effectiveness of the proposed method. It is also worth mention that, as observed from experimental results, incorporating better semantical feature processing module or involving image-level features may further boost the model performance and we leave it to future research.

{\small
\bibliographystyle{ieee}
\bibliography{ms}

\begin{thebibliography}{10}\itemsep=-1pt

\bibitem{bertgit}
BERT.
\newblock https://github.com/google-research/bert.

\bibitem{deeplab}
L.~Chen, G.~Papandreou, I.~Kokkinos, K.~Murphy, and A.~L. Yuille.
\newblock Semantic image segmentation with deep convolutional nets and fully
  connected crfs.
\newblock {\em CoRR}, abs/1412.7062, 2014.

\bibitem{deeplabv1}
L.~Chen, G.~Papandreou, I.~Kokkinos, K.~Murphy, and A.~L. Yuille.
\newblock Deeplab: Semantic image segmentation with deep convolutional nets,
  atrous convolution, and fully connected crfs.
\newblock {\em CoRR}, abs/1606.00915, 2016.

\bibitem{deeplabv3}
L.~Chen, G.~Papandreou, F.~Schroff, and H.~Adam.
\newblock Rethinking atrous convolution for semantic image segmentation.
\newblock {\em CoRR}, abs/1706.05587, 2017.

\bibitem{deeplabv3p}
L.~Chen, Y.~Zhu, G.~Papandreou, F.~Schroff, and H.~Adam.
\newblock Encoder-decoder with atrous separable convolution for semantic image
  segmentation.
\newblock {\em CoRR}, abs/1802.02611, 2018.

\bibitem{5}
A.~Dengel and B.~Klein.
\newblock smartfix: A requirements-driven system for document analysis and
  understanding.
\newblock volume 2423, pages 433--444, 08 2002.

\bibitem{bert}
J.~Devlin, M.~Chang, K.~Lee, and K.~Toutanova.
\newblock {BERT:} pre-training of deep bidirectional transformers for language
  understanding.
\newblock {\em CoRR}, abs/1810.04805, 2018.

\bibitem{7}
D.~Esser, D.~Schuster, K.~Muthmann, and A.~Schill.
\newblock Automatic indexing of scanned documents - a layout-based approach.
\newblock volume 8297, 01 2012.

\bibitem{cloudscan}
R.~B. Palm, O.~Winther, and F.~Laws.
\newblock Cloudscan - {A} configuration-free invoice analysis system using
  recurrent neural networks.
\newblock {\em CoRR}, abs/1708.07403, 2017.

\bibitem{3}
D.~Schuster, K.~Muthmann, D.~Esser, A.~Schill, M.~Berger, C.~Weidling,
  K.~Aliyev, and A.~Hofmeier.
\newblock Intellix - end-user trained information extraction for document
  archiving.
\newblock 08 2013.

\bibitem{hrnet}
K.~Sun, B.~Xiao, D.~Liu, and J.~Wang.
\newblock Deep high-resolution representation learning for human pose
  estimation, 2019.

\bibitem{transformer}
A.~Vaswani, N.~Shazeer, N.~Parmar, J.~Uszkoreit, L.~Jones, A.~N. Gomez,
  L.~Kaiser, and I.~Polosukhin.
\newblock Attention is all you need.
\newblock {\em CoRR}, abs/1706.03762, 2017.

\end{thebibliography}
}

\section{APPENDIX}
\subsection{Model Structure}
We report the CUTIE-B model in Table \ref{tab:cutie}. Tokens are firstly embedd into $128$ dimensional features. Then, $4$ consecutive convolution operations are conducted in the conv block with stride $1$ and $4$ consecutive atrous convolution are conducted in the atrous conv block with stride $1$ and rate $2$. Following the atrous conv block, an ASPP module is employed to fuse multi-resolution features. The low level but high resolution feature from the $1$st output of the convolution block is also added to the model in the shortcut layer with a concatenation operation and a $1\times1$ convolution. Finally, inference output is achieved through a $1\times1$ convolution.
\begin{table*}
	\caption{Structure of the proposed CUTIE-B model.}
\begin{center}
\begin{tabular}{l | c | c | c | c}
	layer name & operations & input dimension & output diemnsion & comments \\
	\hline
	embedding layer & - & 20000 & 128 & \\
	conv block & [$3\times5$]$\times$4 & 256 & 256 & stride=1 \\
	atrous conv block & [$3\times5$]$\times$4 & 256 & 256 & stride=1, rate=2 \\
	ASPP module & [$3\times5$]$\times$3, global pooling, concat, $1\times1$ & 256 & 256 & stride=1, rate=\{4,8,16\} \\
	shorcut layer & concat, $1\times1$ & 256 & 64 & \\
	output layer & $1\times1$ & 64 & 9 & \\ 
\end{tabular}
\end{center}
	\label{tab:cutie}
\end{table*}

\end{document}